\documentclass{article}
\PassOptionsToPackage{numbers, compress}{natbib}

\usepackage[preprint]{neurips_2025}
\usepackage{natbib}

\usepackage{color, colortbl}
\definecolor{Green}{rgb}{0.93,1,0.93}
\definecolor{Gray}{gray}{0.9}
\definecolor{LightBlue}{rgb}{0.88, 0.94, 0.97}
\definecolor{Best}{rgb}{0.85, 0.92, 0.83}
\definecolor{HeaderBlue}{rgb}{0.88, 0.94, 0.97}
\definecolor{GrayRow}{gray}{0.96}
\definecolor{BestRow}{rgb}{0.85, 0.92, 0.83}

\usepackage[utf8]{inputenc} 
\usepackage[T1]{fontenc}    
\usepackage{hyperref}       
\usepackage{url}            
\usepackage{booktabs}       
\usepackage{amsfonts}       
\usepackage{nicefrac}       
\usepackage{microtype}      
\usepackage{xcolor}         
\usepackage{multicol}
\usepackage{multirow}
\usepackage{tabularx}
\usepackage{array}
\usepackage{graphicx}
\usepackage{caption}

\title{Temporal-Oriented Recipe for Transferring Large Vision-Language Model to Video Understanding}

\author{%
    Thong Nguyen$^1$\;\; Zhiyuan Hu$^1$ \; Xu Lin$^1$ \; \\
    \textbf{Cong-Duy Nguyen}$^2$ \; \textbf{See-Kiong Ng}$^1$ \; \textbf{Luu Anh Tuan}$^2$ \\
    $^1$National University of Singapore \;\;  $^2$ Nanyang Technological University 
}

\begin{document}

\maketitle

\begin{abstract}
Recent years have witnessed outstanding advances of large vision-language models (LVLMs). In order to tackle video understanding, most of them depend upon their implicit temporal understanding capacity. As such, they have not deciphered important components that contribute to temporal understanding ability, which might limit the potential of these LVLMs for video understanding. In this work, we conduct a thorough empirical study to demystify crucial components that influence the temporal understanding of LVLMs. Our empirical study reveals that significant impacts are centered around the intermediate interface between the visual encoder and the large language model. Building on these insights, we propose a temporal-oriented recipe that encompasses temporal-oriented training schemes and an upscaled interface. Our final model developed using our recipe significantly enhances previous LVLMs on standard video understanding tasks. \footnote{Our codes and data are available at https://github.com/... (the link is hidden now due to double-blind review)} \end{abstract}

\section{Introduction}
\label{sect:introduction}
Empowered by the elevating popularity of video-text data \citep{nguyen2024encoding, nguyen2024video} and outstanding advances in large language model (LLM)-based designs, recent years have encountered remarkable progress in video understanding with large vision-language models (LVLMs). From the advent of models such as BLIP \citep{li2022blip}, BLIP-2 \citep{li2023blip}, and LLaVA \citep{liu2023visual}, video question answering (VideoQA) has improved from 33.8, 16.7, and 12.4 on MSVD \citep{wu2017deep}, MSRVTT \citep{xu2016msr}, and ActivityNet \citep{krishna2017dense} to more than 60.0 in terms of GPT-3.5 evaluation. Not only VideoQA but also long-term action recognition \citep{kuehne2014language, tang2019coin, wu2021towards} and video captioning \citep{zhou2018towards, islam2024video} have achieved significant breakthroughs.

In recent years, model architectures and training protocols have witnessed significant advancements. However, as these systems grow in diversity and scale, their computational demands pose substantial challenges for comparison, analysis, and reproducibility. Despite these advancements, many approaches have overlooked the core nature of video understanding. Rather than explicitly modeling temporal relationships, they often rely on spatial inductive biases, assuming that spatial knowledge can seamlessly extend to temporal comprehension. For instance, several methods focus on creating a unified representation space for visual and textual modalities \citep{lin2023video, chen2023videollm, zhang2023video}. Others emphasize aggregating or selecting salient visual tokens aligned with prompts \citep{shang2024traveler, xu2024pllava} or leverage large-scale pretraining with instruction-following datasets \citep{maaz2023video, luo2023valley, wang2024gpt4video}. Therefore, existing models fall short of realizing the full potential of video understanding. For example, while VideoQA systems can accurately answer questions about object detection or describe isolated actions, they struggle with queries involving causal and temporal relationships \citep{xiao2021next}. As shown in Table \ref{tab:examples}, they often generate inaccurate responses when faced with questions about temporal order or causality.

\begin{figure*}[t]
    \centering
    \caption{Our temporal-oriented recipe for large vision-language model.}
    \vspace{-5pt}
    \label{fig:recipe}
    \includegraphics[width=0.7\linewidth]{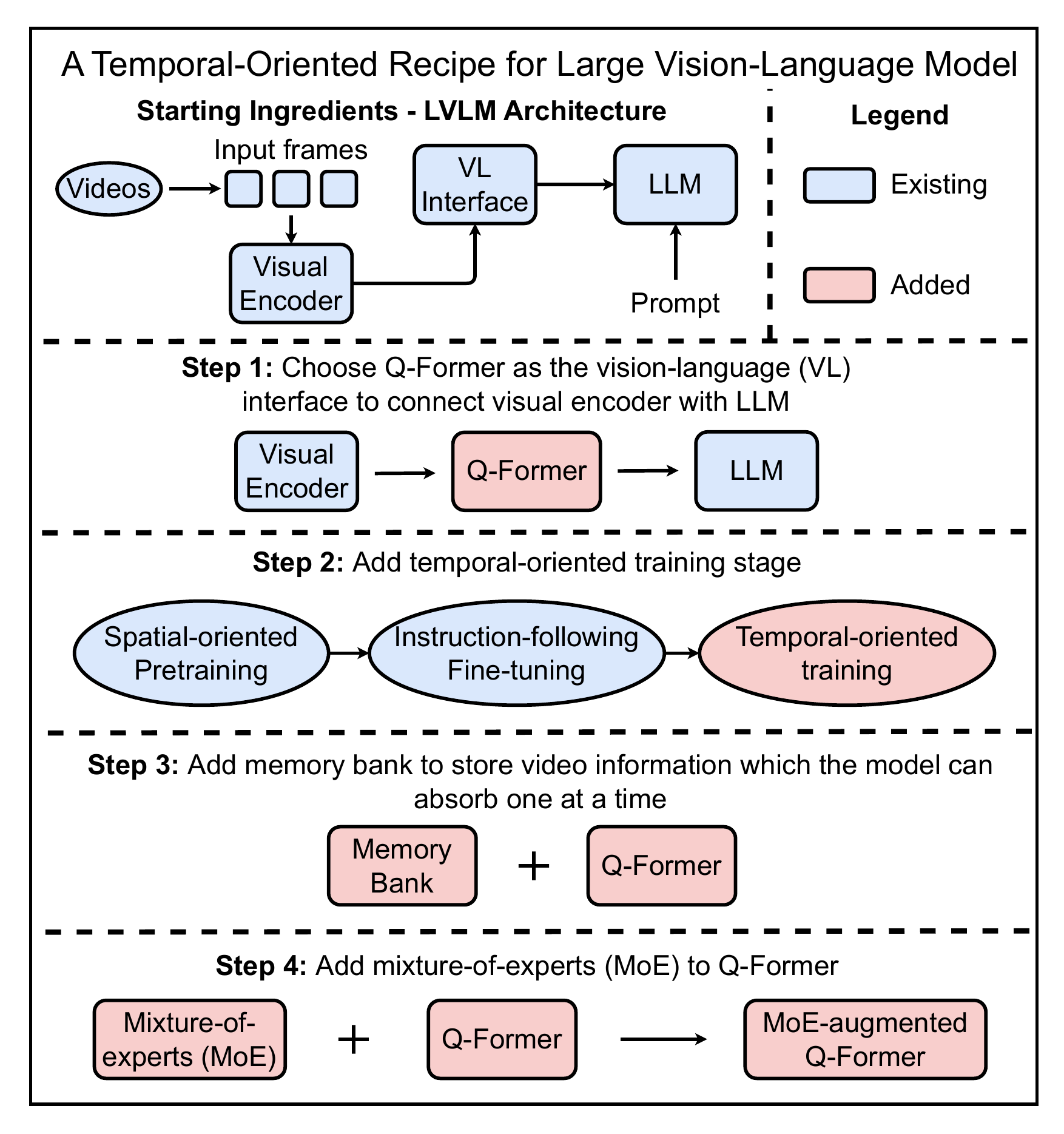}
    \vspace{-25pt}
\end{figure*}

To overcome this limitation, we aim to enhance temporal understanding capabilities of large vision-language models (LVLMs) by advancing temporal-critical components within their architectures. As illustrated in Figure \ref{fig:recipe}, an LVLM is fundamentally composed of three main components: a visual encoder, a vision-language interface, and a large language model (LLM). However, due to the large-scale nature of LLMs and the multimodal complexity of video data, identifying the primary factors driving model effectiveness is challenging \citep{he2024ma, qian2024streaming, chandrasegaran2024hourvideo}, hindering further progress in the field. Our focus is to bridge this gap by ensuring that temporal understanding is treated as a core aspect of video comprehension, rather than an implicit outcome of spatial knowledge.

\begin{table}[t]
    \caption{On the first row, a correct answer should comprise details related to cutting ginger and garlic on a chopping board, whereas other models wrongly mention ``\textit{rub salt}'', ``\textit{cut chicken}'' and ``\textit{add to the pot}'', and ``\textit{pour milk}''. On the second row, we need to respond with ``\textit{getting to the bus}'', but the models mistakenly note ``\textit{late for exam}'', ``\textit{to the hospital}'', and ``\textit{feeling sad}''.}
  \label{tab:examples}
  \vspace{5pt}
  \centering
  \resizebox{0.9\linewidth}{!}{
  \begin{tabular}{@{} l | p{0.18\linewidth}  |p{0.18\linewidth} | p{0.18\linewidth}  |p{0.18\linewidth} | p{0.18\linewidth} @{} }
    \toprule
    \rowcolor{gray!30}
    \multicolumn{1}{c|}{\textbf{Video}}        & \multicolumn{1}{c|}{\textbf{Question}}    & \multicolumn{1}{c|}{\textbf{Sample Answer}}  & \multicolumn{1}{c|}{\textbf{Video-LLaMA}~\citep{zhang2023video}} & \multicolumn{1}{c|}{\textbf{Video-LLaVA}~\citep{lin2023video}} & \multicolumn{1}{c|}{\textbf{Qwen2.5-VL}~\citep{bai2025qwen2}} \\
    \midrule
    \multirow{3}{*}{%
      \includegraphics[width=0.09\linewidth]{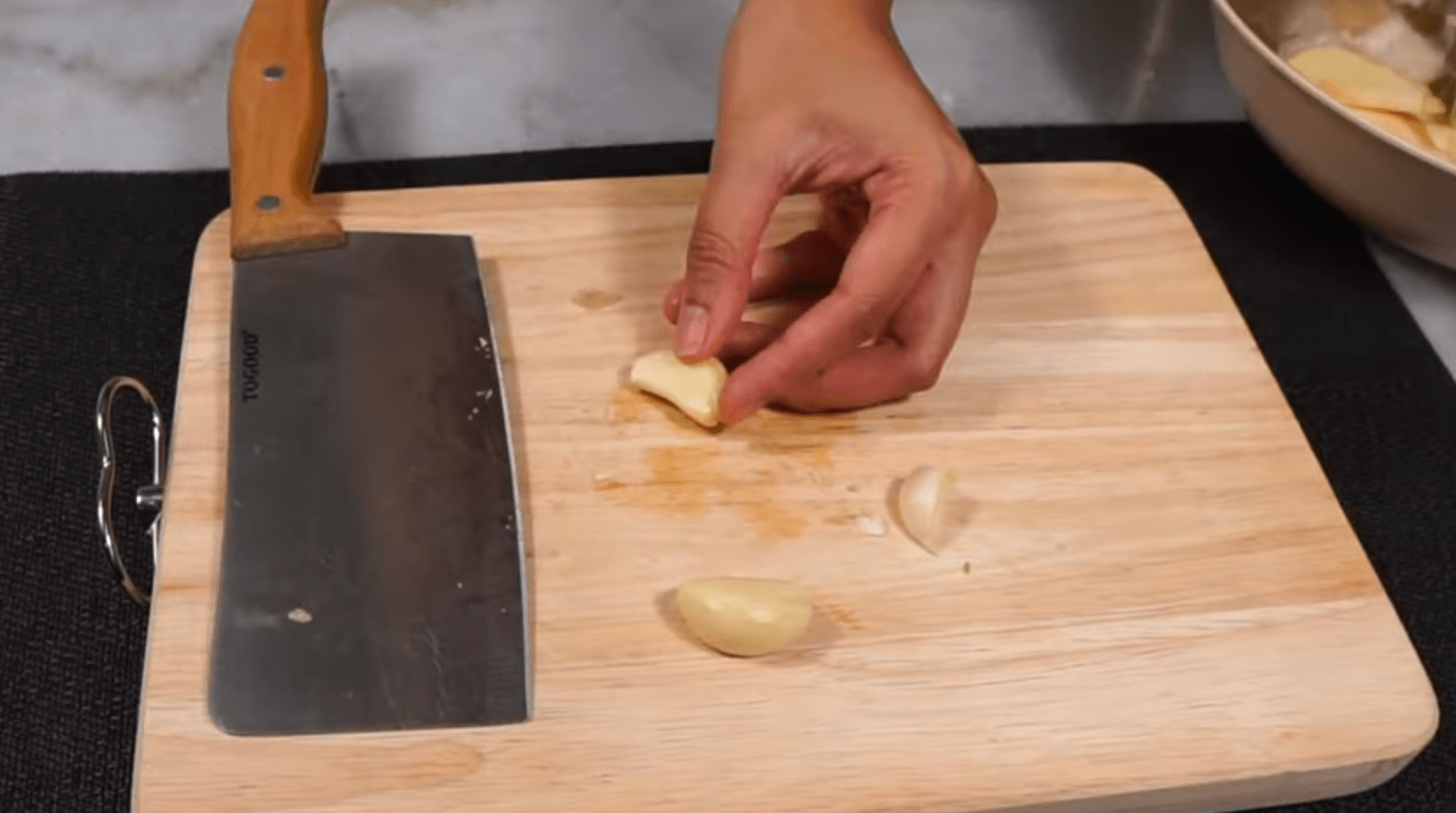}\quad
      \includegraphics[width=0.09\linewidth]{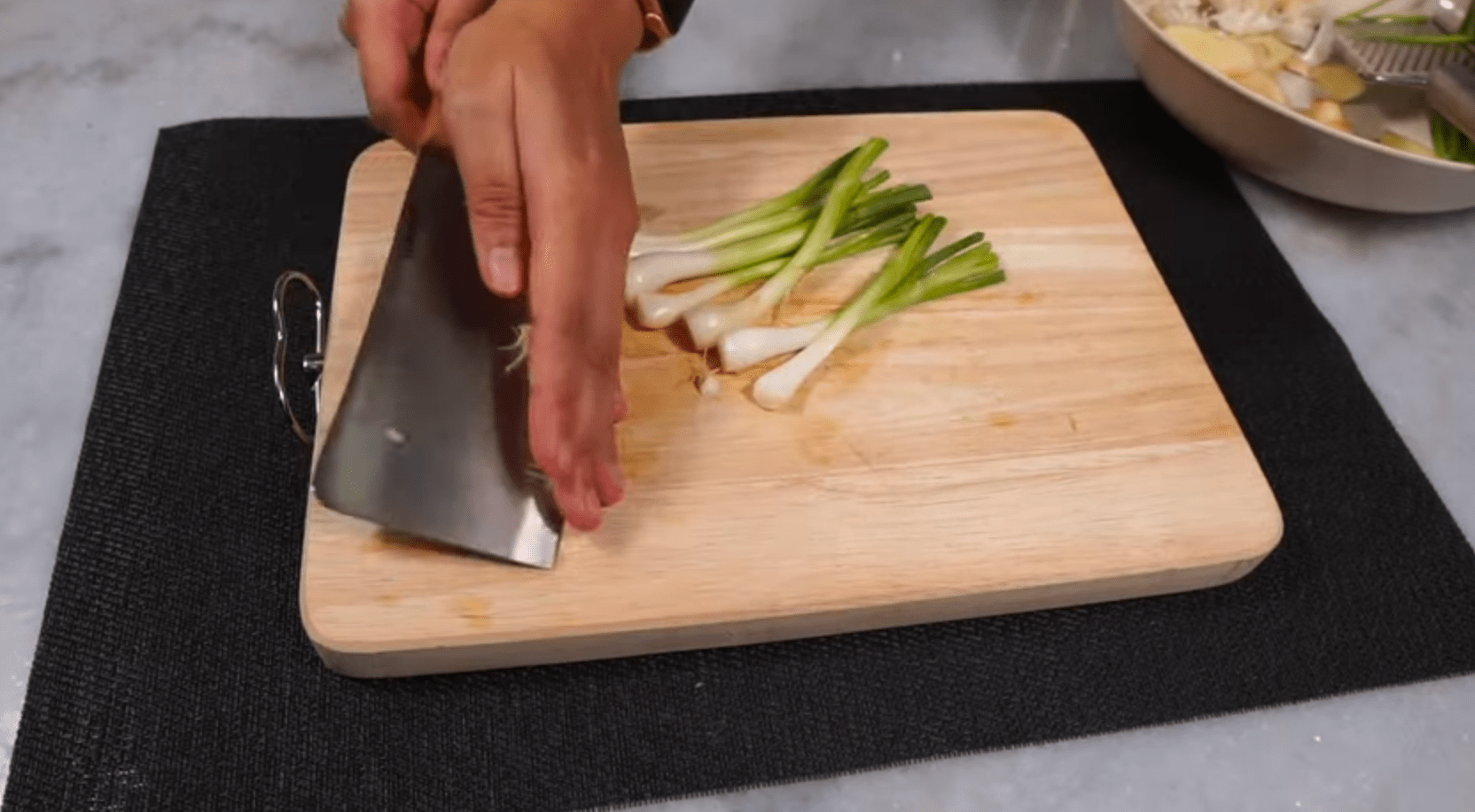}\quad
      \includegraphics[width=0.09\linewidth]{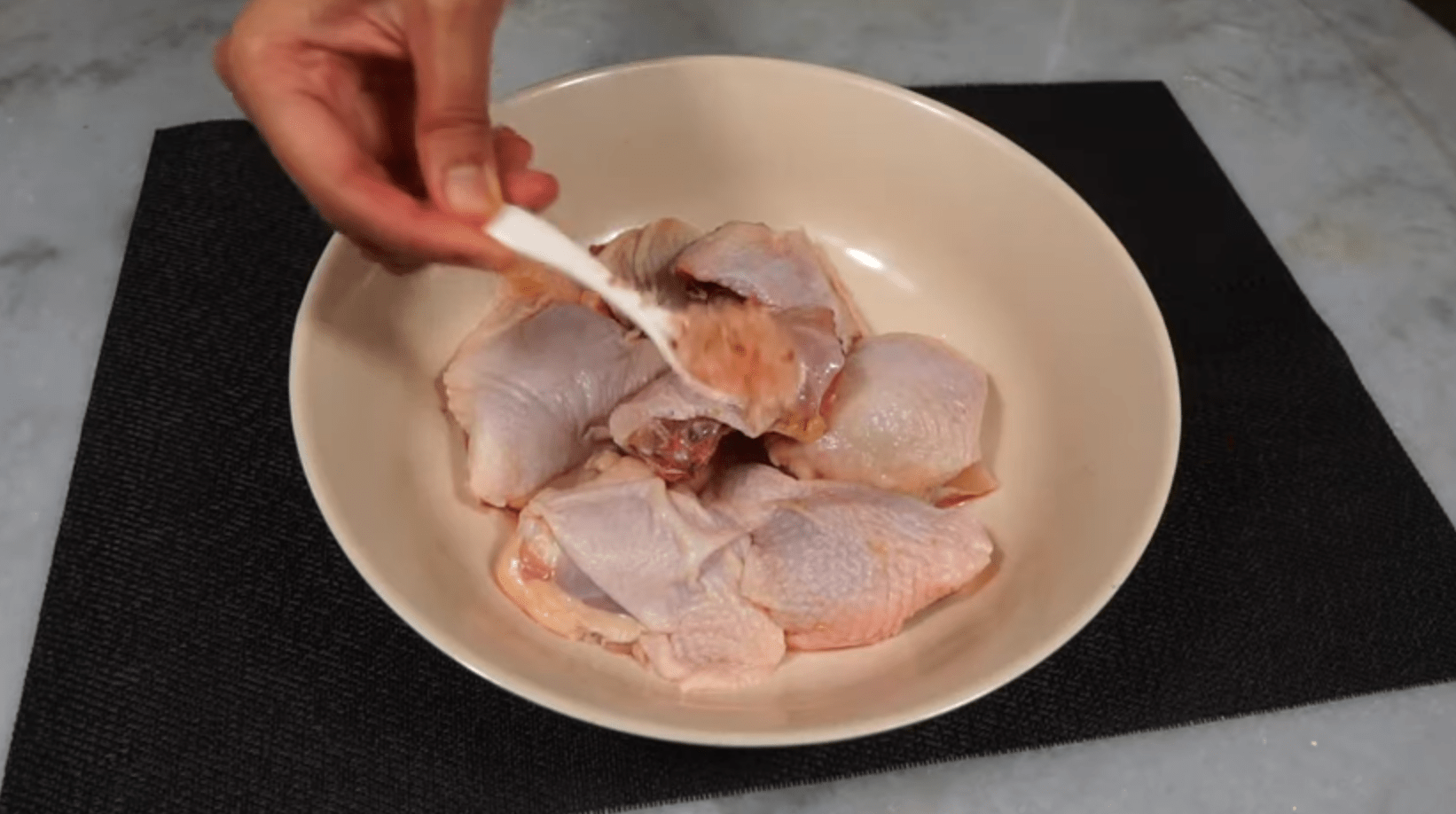}\quad
      \includegraphics[width=0.09\linewidth]{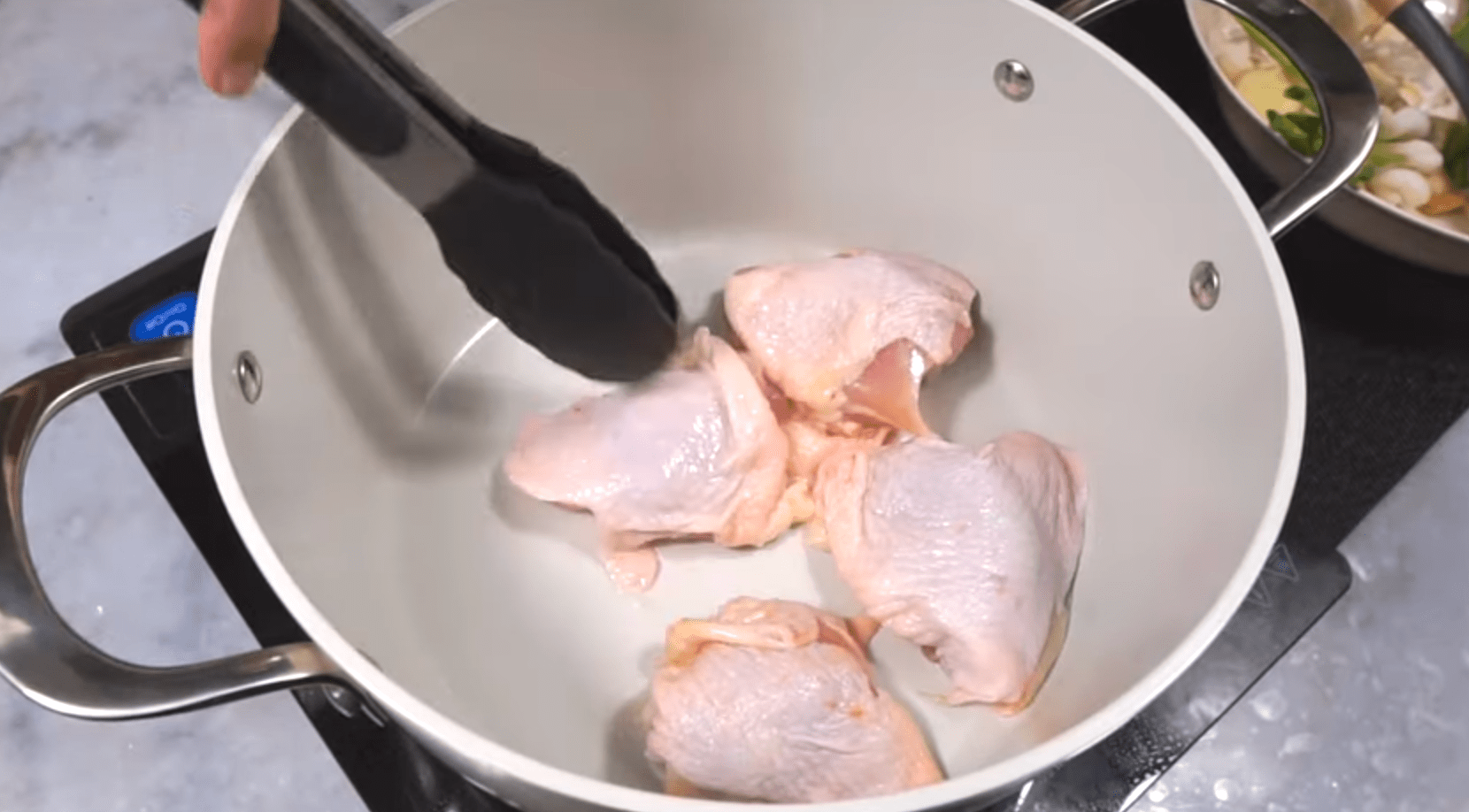}}  
      & What does the person do before seasoning chicken?  
      & The person cuts ginger and garlic into small pieces on a chopping board  
      & The person rubs salt on the chicken.  
      & Before seasoning chicken, the person cuts it into small pieces and adds it to a pot.  
      & Before seasoning chicken, the person pours milk into a bowl of rice.  \\ \hline
    \addlinespace[3pt]
    \multirow{3}{*}{%
      \includegraphics[width=0.09\linewidth]{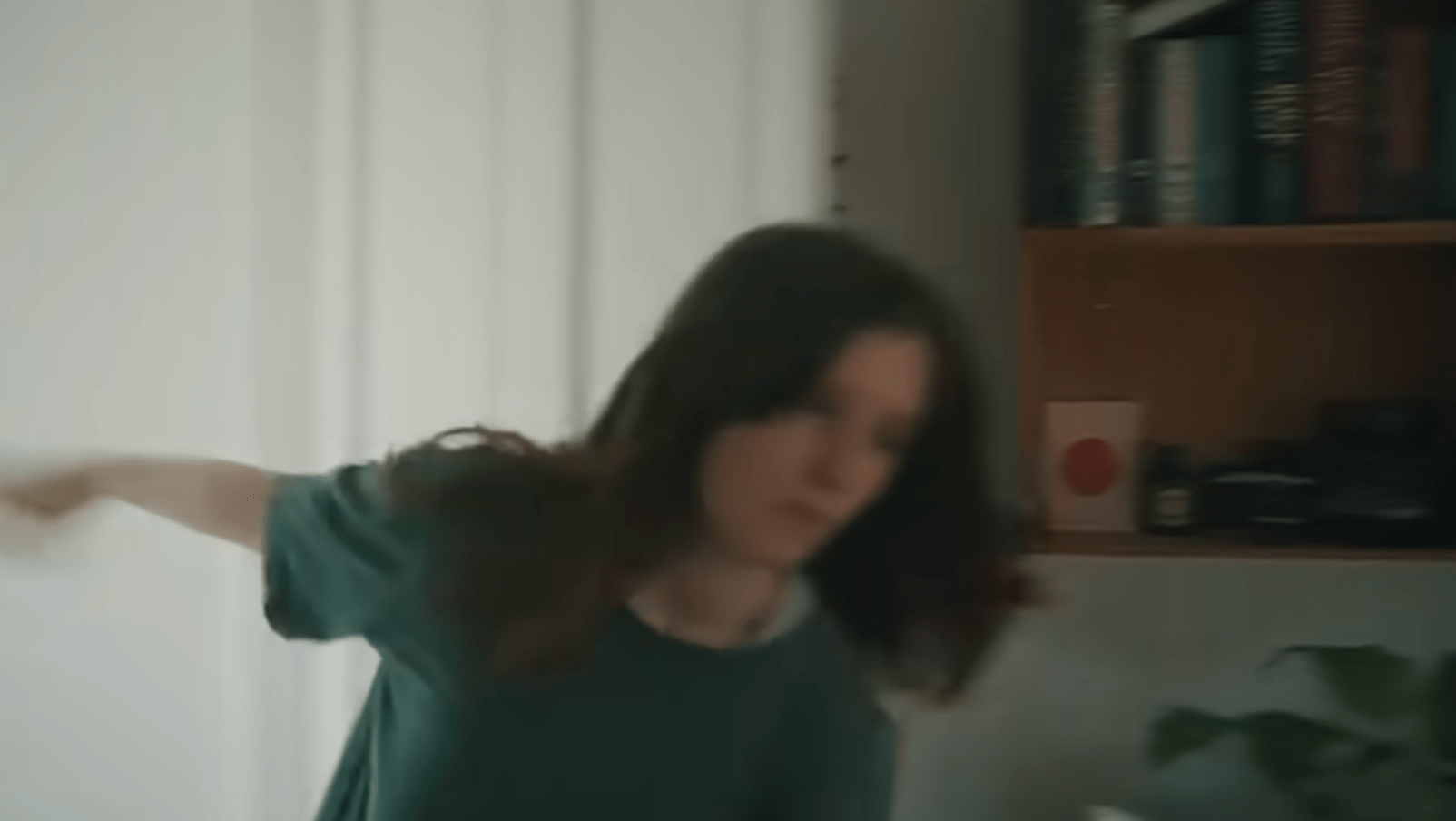}\quad
      \includegraphics[width=0.09\linewidth]{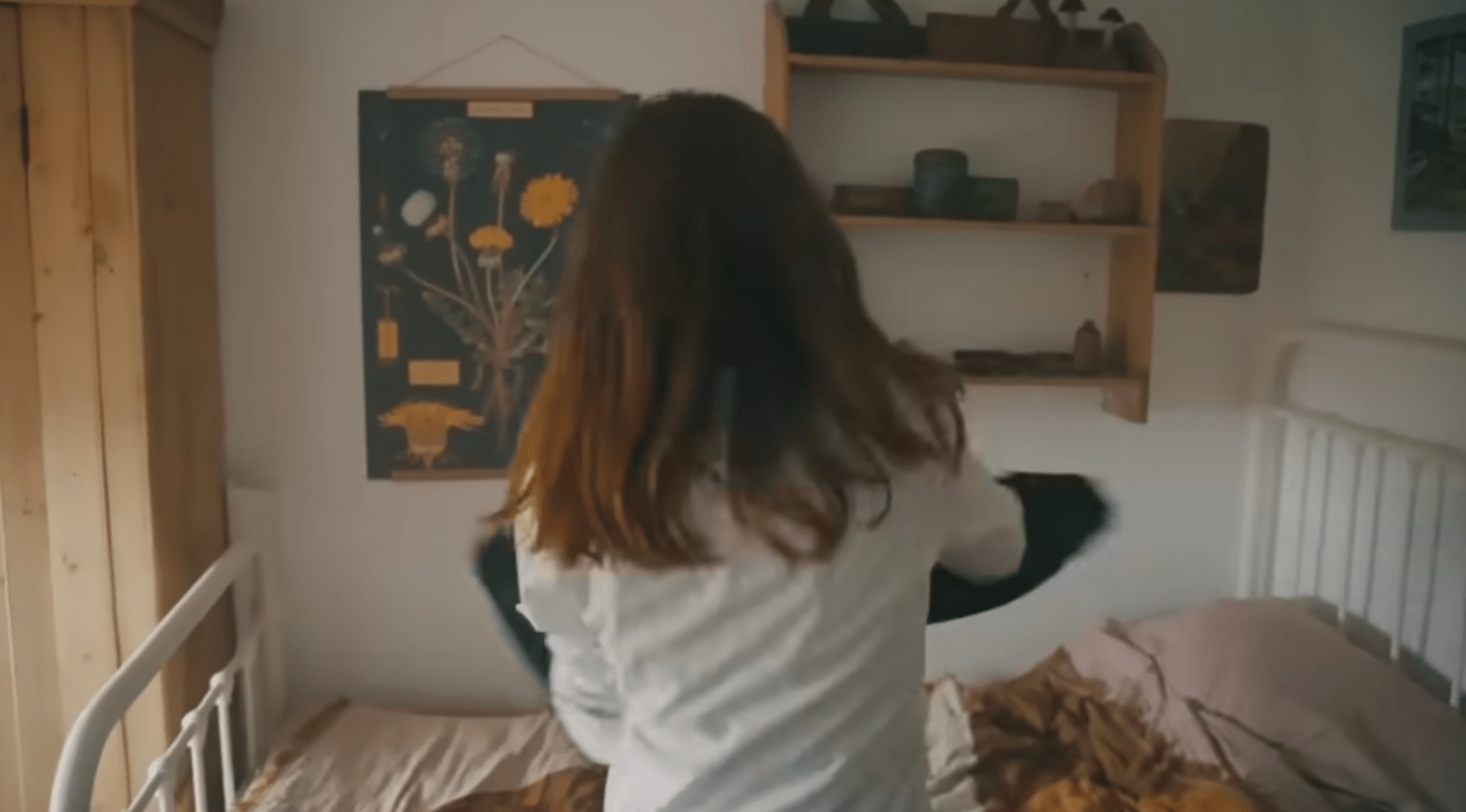}\quad
      \includegraphics[width=0.09\linewidth]{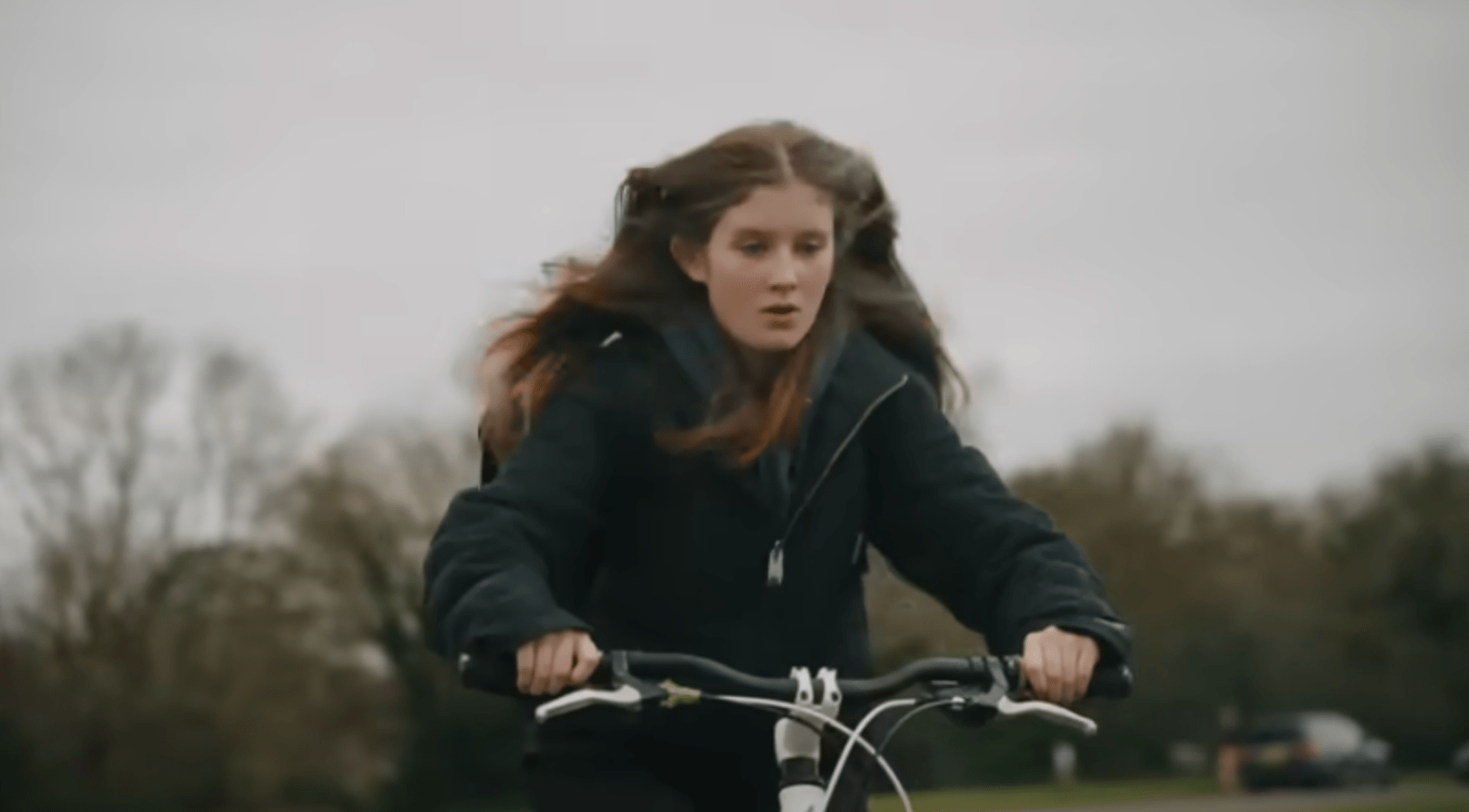}\quad
      \includegraphics[width=0.09\linewidth]{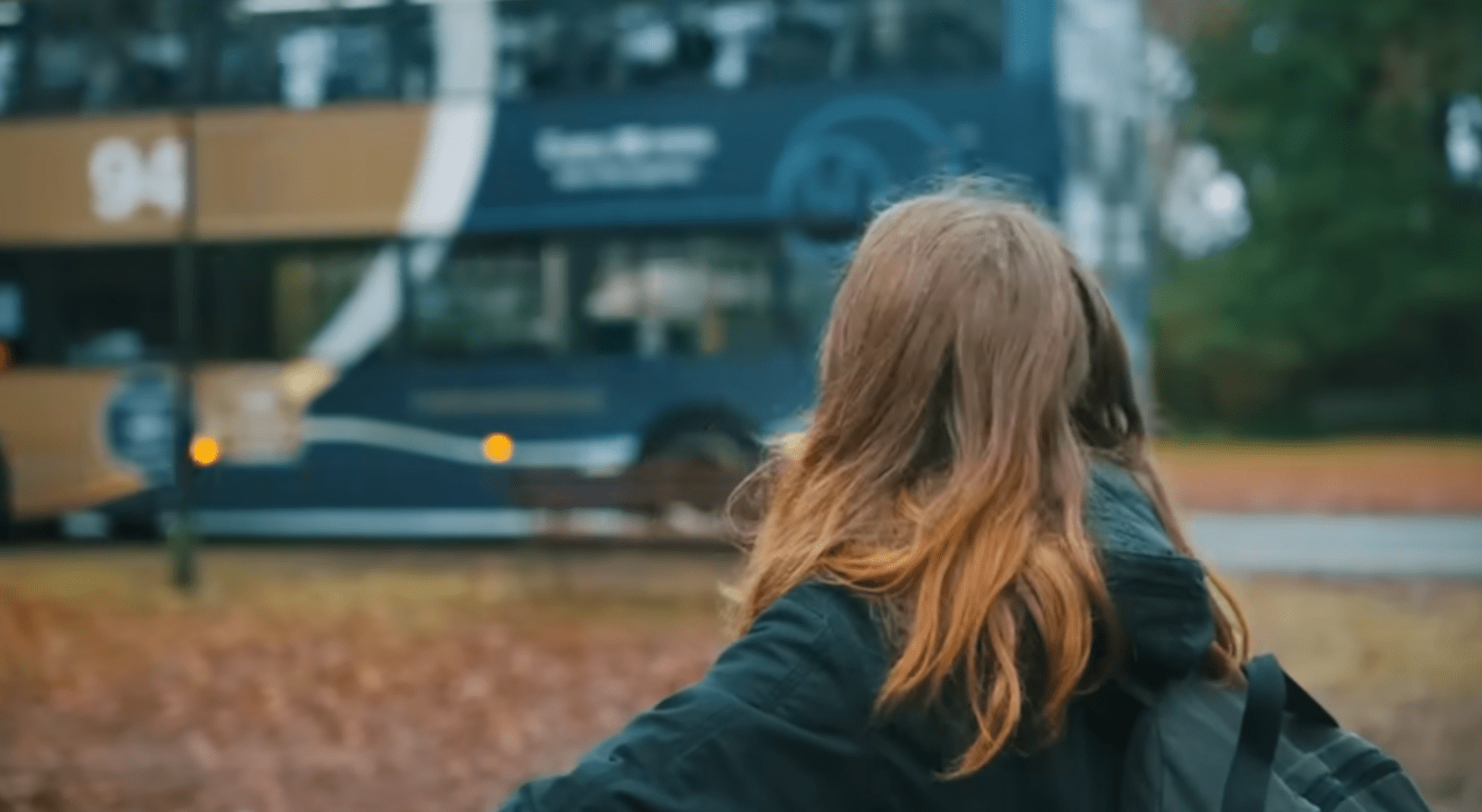}}  
      & Why did the girl rush?   
      & The girl rushed to get to the bus on time, as she was running to catch the bus  
      & The girl rushed because she was late for her exam.  
      & The girl rushed to the hospital because she was feeling unwell.  
      & The girl in the picture appears to be looking down and might be feeling sad or contemplative.  \\
    \bottomrule
  \end{tabular}}
  \vspace{-15pt}
\end{table}

To illustrate our points, in Table \ref{tab:lvlm_models}, we explicate the diversity of modern LVLMs for video understanding by examining them along various dimensions, including visual encoder, vision-language interface, LLM, and training data. Based on this examination, we observe that there exist stark differences among these models. Nevertheless, it is not straightforward to dissect which factors make an important contribution to the overall video understanding performance and which do not. 

{\renewcommand{\arraystretch}{1.2}
\begin{table}[t]
\caption{Existing LVLM models exhibit stark distinctions among themselves, making it challenging to reproduce, analyze, and compare these methods. Therefore, we aim to answer the question: ``\textit{Is there a straightforward recipe to build temporal understanding capacity for LVLMs?}''}
\vspace{5pt}
\label{tab:lvlm_models}
\centering
\resizebox{0.95\linewidth}{!}{
\begin{tabular}{l|c|l|c|c|c}
\toprule
\rowcolor{HeaderBlue}
\multicolumn{1}{c|}{\textbf{Method}} & \textbf{Visual Encoder} & \multicolumn{1}{c|}{\textbf{Vision-Language Interface}} & \textbf{LLM Size} & \textbf{Training Data Size (Pretraining)} & \textbf{Training Data Size (IFT)} \\ 
\midrule
VALLEY \citep{luo2023valley}         & ViT-L        & Transformer + Mean-pooling                       & 13B      & 702K                 & 73K  \\
\rowcolor{GrayRow}
Video-LLaMA \citep{zhang2023video}   & CLIP-G       & Q-Former                                         & 7B       & 3M                   & 18K  \\
LLaMA-VID \citep{li2024llama}        & CLIP-G       & Linear Projection                                & 13B      & 790K                 & 763K \\
\rowcolor{GrayRow}
VideoChat \citep{li2023videochat}    & CLIP-G       & Q-Former                                         & 7B       & 25M                  & 18K  \\
VideoChat2 \citep{li2024mvbench}     & UMT-L        & Q-Former                                         & 7B       & 25M                  & 2M   \\
\rowcolor{GrayRow}
Video-ChatGPT \citep{maaz2023video}  & ViT-L        & Mean-pooling + Linear Projection                  & 7B       & 595K                 & 100K \\
Video-LLaVA \citep{lin2023video}     & ViT-L        & Linear Projection                                & 7B       & 1.3M                 & 765K \\
\rowcolor{GrayRow}
GPT4Video \citep{wang2024gpt4video}  & ViT-L        & Q-Former                                         & 7B       & 11K                  & 50K  \\
PLLaVA \citep{xu2024pllava}          & ViT-L        & Linear Projection + Adaptive Pooling             & 7B/13B/34B & 25M              & 783K \\
\rowcolor{GrayRow}
ST-LLM \citep{liu2024st}             & BLIP-2       & Linear Projection                                & 7B       & 25M                  & 2M   \\
Chat-UniVi \citep{jin2024chat}       & ViT-L        & Clustering-based Merging + Linear Projection      & 7B/13B   & 1.6M                 & 649K \\
\bottomrule
\end{tabular}}
\vspace{-15pt}
\end{table}}

As one of the works that initiates empirical analysis research line, METER \citep{dou2022empirical} studies a wide variety of components in the context of image-language modeling. Unfortunately, its analysis mostly works on images and neglects many aspects related to video modeling, such as spatio-temporal architectural design, video pretraining data, and video pretraining objectives. To fill in such gap, recent VindLU work \citep{cheng2023vindlu} conducts an analysis towards important factors for video-language understanding models. Unfortunately, their analysis is limited to small-scale frameworks with millions of parameters. Similarly, \citet{fu2023empirical} performs an empirical study of video-language transformers, but narrowly concentrates on masked visual modeling objective. 

Our primary objective in this work is to answer the question ``\textit{Is there a straightforward recipe to build temporal understanding capacity for LVLMs?}'' Our answer is \textbf{yes}. To arrive at the answer, we conduct a thorough empirical study that demystifies the importance of various design choices and ultimately leads to a temporal-oriented recipe that significantly enhances video understanding results of previous LVLMs. Our recipe starts from a standard paradigm of a large vision-language model then proceeds with a progressive expansion scheme, where at each stage, we investigate a specific aspect of LVLM framework design (\textit{e.g.}, architecture, training objective, training data, etc.) and choose the most effective option. Particularly, we study the following LVLM design components: (i) the vision-language interface, (ii) the video training protocols, and (iii) temporal memory bank, and (iv) scaling of the essential component. We present our recipe in Figure \ref{fig:recipe}. 

The key lessons of our study include:
\vspace{-10pt}
\begin{itemize}
    \itemsep -1pt
    \item Among components in an LVLM architecture, we discover that enhancing vision-language interface significantly advances the temporal modeling strength of the LVLM. 
    \item A query transformer that incorporates query tokens to interact with video representations combined with a temporal memory bank to compress salient video information is crucial for satisfactory video understanding performance.
    \item We can further obtain gains of temporal understanding level with techniques to scale up the interface, including mixture-of-experts and number of query tokens to store video information.
    \item An additional training stage for LVLMs with temporal-oriented data is sufficient to remarkably enhance temporal understanding capability and achieve impressive results on video understanding.
\end{itemize}
\vspace{-10pt}

\vspace{-5pt}
\section{Related Work}
\vspace{-5pt}
\label{sect:related_work}
\vspace{-5pt}
\subsection{Large Vision-Language Model}
Recent advancements in large language model (LLM)-based models have led to the development of powerful large vision-language models (LVLMs) with visual understanding capabilities. Despite these advances, even state-of-the-art models---such as VALLEY \citep{luo2023valley}, Video-LLaMA \citep{zhang2023video}, LLaMA-VID \citep{li2024llama}, VideoChat \citep{li2023videochat}, VideoChat2 \citep{li2024mvbench}, Video-ChatGPT \citep{maaz2023video}, Video-LLaVA \citep{lin2023video}, GPT4Video \citep{wang2024gpt4video}, PLLaVA \citep{xu2024pllava}, ST-LLM \citep{liu2024st}, and Chat-UniVi \citep{jin2024chat}---typically follow a common architectural paradigm comprising a visual encoder, a vision-language interface, and LLM. However, significant differences among these implementations obscure the key contributing factors to their performance, making it challenging to adapt LVLMs effectively to video understanding. Efforts to demystify critical components, such as VindLU \citep{cheng2023vindlu} and METER \citep{dou2022empirical}, provide valuable insights. Unfortunately, VindLU \citep{cheng2023vindlu} focuses primarily on medium-scale transformer-based models with millions of parameters, while METER \citep{dou2022empirical} is limited to image-based analysis, leaving its applicability to video data uncertain. In contrast, our work targets large-scale models with billions of parameters that are specifically designed for video understanding, aiming to provide clearer guidance for future research in this domain.

\subsection{Video Understanding}
\vspace{-5pt}
In recent years, large-scale video understanding has made significant strides \citep{li2023videochat, li2024mvbench, nguyen2025multi, nguyen2025motion, chen2023videollm, lin2023video}. Recent studies have shown impressive performance across a range of downstream tasks, including video captioning \citep{chen2023videollm, wang2024grounded} and video question answering \citep{lin2023video, maaz2023video, li2024llama}. Many LVLMs \citep{zhang2023video, li2024llama, lin2023video, xu2024pllava} have achieved notable success by connecting visual encoder and the LLM using a vision-language interface, which is popularly implemented as a linear projection layer or a Q-Former. To train these LVLMs, several works utilize pretraining and instruction-tuning datasets, often comprising a joint mixture of image- and video-text pairs. However, this mixture complicates the disentanglement of spatial and temporal understanding, making it unclear which components are essential for enabling temporal reasoning in LVLMs. In contrast to the image understanding domain, there remains a lack of empirical studies that systematically investigate the design choices and foundational components necessary for video understanding with LVLMs.
\vspace{-5pt}
\section{Temporal-Oriented Recipe for Large Vision-Language Model}
\vspace{-5pt}
\label{sect:recipe}
In this section, we delineate our temporal-oriented recipe for large vision-language model. We start with a standard large vision-language model (LVLM), which consists of a visual encoder such as ViT and a large language model (LLM). Then, we progressively expand it to a model that achieves impressive temporal understanding results on various video understanding datasets and tasks. At each step of our recipe, we investigate how design choices have an impact upon temporal capacity of the LVLM. Throughout our procedure, we will discover answers to the following questions about the temporal-oriented recipe design:
\begin{itemize}
    \item Can explicitly constructing temporal understanding capacity help LVLM, particularly provided that various video understanding benchmarks are spatially biased \citep{buch2022revisiting, lei2022revealing}? If so, what is the best mechanism for LVLM to conduct temporal modeling?
    \item Given that video lengths vary with a wide range, what is the most productive mechanism for LVLM to read/absorb video information? Several approaches use visual encoder combined with linear projection \citep{maaz2023video, xu2024pllava} or query transformer (Q-Former) \citep{li2023videochat, li2024mvbench}, then proceed with a pooling mechanism, whereas others adopt a memory bank \citep{he2024ma}. Which of these is the most effective?
    \item Which temporal-oriented training schemes are most useful for temporal representation learning? There exist a wide variety of schemes, including video captioning \citep{abdar2024review, abdar2024review}, moment captioning \citep{qasim2025dense, yang2023vid2seq}, moment grounding \citep{nguyen2023demaformer, lei2021detecting}, and video summarization \citep{apostolidis2021video, nguyen2023read}. How significant is each of these schemes? Are they complementary to each other?
    \item How can we optimize temporal capacity of LVLM? Can we inherit the mixture-of-experts (MoE) approach from LLM works, or increase the number of query tokens?
\end{itemize}

\vspace{-5pt}
\subsection*{Step 0: Starting Ingredients}
\noindent\textbf{Large Vision-Language Model.} We start with a standard ViT-G/14 \citep{dosovitskiy2020image} from EVA-CLIP \citep{fang2023eva}. For LLM, we use either Vicuna-7B or Vicuna-13B \citep{chiang2023vicuna}, forming either a 7B-LVLM or a 13B-LVLM, respectively. Formally, given a paired video and text prompt $(V, T)$, the visual encoder randomly selects a sequence of frames from the video as input to extract visual embeddings. The LLM encodes the prompt $T$ to extract the textual embeddings. 

\noindent\textbf{Experimental Setup.} As our initialization, we directly inherit the pretrained and instruction-tuned model on image-based data \citep{liu2023visual, li2022blip, li2023blip}. Afterwards, we either conduct an additional temporal-oriented training step or go straight to finetuning and evaluating the model on the seven popular video understanding datasets: MSRVTT \citep{xu2016msr}, MSVD \citep{chen2011collecting}, ActivityNet-QA \citep{caba2015activitynet}, Breakfast \citep{kuehne2014language}, COIN \citep{tang2019coin}, and LVU \citep{wu2021towards}. For our empirical investigation, we choose the video question answering (VideoQA) task and report the accuracy across these datasets.

In the following subsections, we progressively expand this baseline by adding more components of elevating complexity. Specifically, we start by incorporating vision-language interface (step 1), integrate a temporal-oriented training stage (step 2), insert a temporal memory bank (step 3), and upscaling the interface (step 4). Note that due to the large computational cost, we cannot ablate the order of the steps in our recipe. Therefore, the order of the steps is primarily determined by the computational cost (\textit{i.e.} the steps that can be implemented most efficiently are investigated before other steps, subsequently moving to more computationally costly steps).

\vspace{-5pt}
\subsection*{Step 1: Vision-Language Interface for LVLM}

{\renewcommand{\arraystretch}{1.2}
\begin{table}[t]
\caption{Effect of different types of vision-language interface on 7B-LVLM}
\label{tab:step1_vision_language_interface_7b}
\centering
\resizebox{0.9\linewidth}{!}{
\begin{tabular}{l|c|c|c|c|c|c}
\toprule
\rowcolor{LightBlue}
\multicolumn{1}{c|}{\textbf{Vision-Language Interface}} & \textbf{MSRVTT} & \textbf{MSVD} & \textbf{ActivityNet-QA} & \textbf{Breakfast} & \textbf{COIN} & \textbf{LVU}  \\
\midrule
Linear Projection                                      & 45.3            & 56.9          & 46.3                    & 85.9               & 83.9          & 57.1         \\
\rowcolor{Gray}
Q-Former w/o SA + Mean-pooling - S = 3                 & 45.8            & 57.4          & 46.4                    & 86.3               & 84.8          & 58.0         \\
Q-Former w/o SA + Adaptive-pooling - S = 3             & 46.2            & 57.5          & 46.8                    & 87.0               & 85.6          & 58.2         \\
\rowcolor{Gray}
Q-Former w/o SA + ESA - S = 3                          & 46.3            & 57.7          & 47.2                    & 87.6               & 86.0          & 58.7         \\
Q-Former w/o SA + Mean-pooling - S = 6                 & 46.3            & 58.1          & 47.6                    & 87.9               & 86.5          & 58.8         \\
\rowcolor{Gray}
Q-Former w/o SA + Adaptive-pooling - S = 6             & 46.6            & 58.1          & 47.6                    & 88.3               & 87.1          & 59.2         \\
Q-Former w/o SA + ESA - S = 6                          & 47.0            & 58.2          & 48.1                    & 88.4               & 87.6          & 59.8         \\
\rowcolor{Gray}
Q-Former w/o SA + Mean-pooling - S = 9                 & 47.2            & 58.2          & 48.3                    & 88.5               & 87.9          & 60.3         \\
Q-Former w/o SA + Adaptive-pooling - S = 9             & 47.4            & 58.7          & 48.6                    & 89.1               & 88.5          & 60.8         \\
\rowcolor{Gray}
Q-Former w/o SA + ESA - S = 9                          & 47.5            & 59.1          & 48.8                    & 89.9               & 88.6          & 61.2         \\
Q-Former w/o SA + Mean-pooling - S = 12                & 47.6            & 59.2          & 49.3                    & 90.1               & 89.1          & 61.8         \\
\rowcolor{Gray}
Q-Former w/o SA + Adaptive-pooling - S = 12            & 47.9            & 59.8          & 49.5                    & 90.8               & 89.4          & 62.2         \\
Q-Former w/o SA + ESA - S = 12                         & 48.2            & 59.9          & 49.5                    & 90.9               & 90.0          & 62.4         \\
\rowcolor{Gray}
Q-Former w/ SA - S = 3                                 & 48.7            & 60.2          & 49.8                    & 91.7               & 90.8          & 62.7         \\
Q-Former w/ SA - S = 6                                 & 48.8            & 60.4          & 49.8                    & 91.9               & 91.6          & 52.0         \\
\rowcolor{Gray}
Q-Former w/ SA - S = 9                                 & 49.0            & 60.4          & 50.3                    & 92.1               & 92.5          & 63.5         \\
Q-Former w/ SA - S = 12                                & 49.3            & 60.4          & 50.3                    & 92.4               & 92.7          & 63.7         \\
\rowcolor{Best}
Pre-trained Q-Former w/ SA - S = 12                    & \textbf{49.4}            & \textbf{60.8}          & \textbf{50.6}                    & \textbf{93.1}               & \textbf{93.4}          & \textbf{63.8}    \\
\bottomrule
\end{tabular}}
\vspace{-10pt}
\end{table}}

{\renewcommand{\arraystretch}{1.2}
\begin{table}[t]
\caption{Effect of different types of vision-language interface on 13B-LVLM}
\label{tab:step1_vision_language_interface_13b}
\centering
\resizebox{0.9\linewidth}{!}{
\begin{tabular}{l|c|c|c|c|c|c}
\toprule
\rowcolor{LightBlue}
\multicolumn{1}{c|}{\textbf{Vision-Language Interface}} & \textbf{MSRVTT} & \textbf{MSVD} & \textbf{ActivityNet-QA} & \textbf{Breakfast} & \textbf{COIN} & \textbf{LVU}  \\
\midrule
Linear Projection                                      & 56.2            & 67.9          & 48.7                    & 86.5               & 84.5          & 65.2         \\
\rowcolor{Gray}
Pre-trained Q-Former w/ SA - S = 12                    & 56.8            & 68.2          & 48.7                    & 86.6               & 85.1          & 67.2         \\
Q-Former w/o SA + Mean-pooling - S = 3                 & 56.9            & 68.3          & 49.0                    & 87.2               & 85.2          & 67.3         \\
\rowcolor{Gray}
Q-Former w/o SA + Adaptive-pooling - S = 3             & 57.1            & 68.7          & 49.3                    & 87.7               & 86.0          & 67.6         \\
Q-Former w/o SA + ESA - S = 3                          & 57.5            & 69.3          & 49.5                    & 88.0               & 86.0          & 68.1         \\
\rowcolor{Gray}
Q-Former w/o SA + Mean-pooling - S = 6                 & 57.8            & 69.7          & 49.5                    & 88.4               & 86.4          & 68.1         \\
Q-Former w/o SA + Adaptive-pooling - S = 6             & 58.0            & 69.8          & 49.8                    & 88.7               & 87.3          & 68.5         \\
\rowcolor{Gray}
Q-Former w/o SA + ESA - S = 6                          & 58.1            & 70.4          & 50.0                    & 88.8               & 87.8          & 69.0         \\
Q-Former w/o SA + Mean-pooling - S = 9                 & 58.2            & 71.0          & 50.2                    & 89.3               & 88.2          & 69.1         \\
\rowcolor{Gray}
Q-Former w/o SA + Adaptive-pooling - S = 9             & 58.3            & 71.2          & 50.3                    & 89.4               & 88.9          & 69.4         \\
Q-Former w/o SA + ESA - S = 9                          & 58.7            & 72.8          & 50.4                    & 90.2               & 89.5          & 69.6         \\
\rowcolor{Gray}
Q-Former w/o SA + Mean-pooling - S = 12                & 59.1            & 72.2          & 50.5                    & 90.6               & 90.2          & 69.8         \\
Q-Former w/o SA + Adaptive-pooling - S = 12            & 59.2            & 72.3          & 50.6                    & 91.3               & 90.7          & 70.1         \\
\rowcolor{Gray}
Q-Former w/o SA + ESA - S = 12                         & 59.5            & 72.5          & 51.1                    & 92.1               & 90.9          & 70.3         \\
Q-Former w/ SA - S = 3                                 & 59.8            & 72.5          & 51.5                    & 92.4               & 91.1          & 71.0         \\
\rowcolor{Gray}
Q-Former w/ SA - S = 6                                 & 59.9            & 73.1          & 51.7                    & 92.4               & 91.5          & 71.2         \\
Q-Former w/ SA - S = 9                                 & 60.3            & 73.8          & 51.8                    & 92.9               & 92.0          & 71.5         \\
\rowcolor{Gray}
Q-Former w/ SA - S = 12                                & 60.5            & 74.3          & 52.0                    & 93.7               & 92.4          & 71.5         \\
\rowcolor{Best}
Pre-trained Q-Former w/o SA - S = 12                   & \textbf{60.6}            & \textbf{74.3}          & \textbf{52.5}                    & \textbf{93.7}               & \textbf{93.2}          & \textbf{71.8}     \\
\bottomrule
\end{tabular}}
\vspace{-10pt}
\end{table}}

In the first stage of our temporal-oriented recipe, we investigate the interface between the vision and language domain for our LVLM. Such interface will enable the LLM to have access to visual information from the video input. For compactness, we study three interface schemes:

\begin{itemize}
    \item \textbf{Linear projection:} In this interface, the linear projection maps visual embeddings into appropriate dimensional space for the LLM. Due to its simplicity, this approach has been widely adopted by previous LLaVA-based LVLMs \citep{xu2024pllava, lin2023video}.

    \item \textbf{Query Transformer with Self-Attention (Q-Former w/ SA):} Following \citep{zhang2023video, he2024ma}, we use a number of transformer submodules which consist of cross-attention and self-attention layers. Cross-attention layers will enable a set of learnable query embeddings to interact with video representations to extract video information. In this variant, our Q-Former also contains self-attention layers, which can perform temporal modeling since they relate video frames together. We vary the number of submodules $S \in \{3,6,9,12\}$. Parameters of Q-Former can be either randomly initialized or initialized from a pre-trained model. In our work, if we initialize Q-Former from a pre-trained model, we follow MA-LMM \citep{he2024ma} to use the \textit{bert-base-uncased} with $S = 12$ submodules.

    \item \textbf{Query Transformer without Self-Attention (Q-Former w/o SA):} This version is similar to the previous one, except the fact that Q-Former does not comprise self-attention layers. Therefore, we need to incorporate an additional component after Q-Former for temporal modeling. We experiment with possible choices, including mean-pooling, adaptive pooling, and external self-attention (ESA) layers. 
\end{itemize}

As Table \ref{tab:step1_vision_language_interface_7b} and \ref{tab:step1_vision_language_interface_13b} show, Q-Former demonstrates critical performance improvement over the linear projection approach. The improvement is indicated by average +6.0\% and +6.2\% accuracy boost of our 12-layer Q-Former variant over the 7B and 13B linear-projection baseline, respectively. We also observe that initializing Q-Former self-attention layers with pretrained BERT encoder makes a significant contribution to the performance boost. This suggests that temporal semantics among words can be related to temporal relations among video frames. 

Interestingly, our findings contradict the conclusions of several prior studies \citep{liu2023visual, koh2023grounding}, which suggest that a simple linear projection is sufficient---and even more effective---than the Q-Former approach. In contrast, we observe that Q-Former plays a crucial role due to its ability to model diverse temporal relations across a broad range of video scenarios. We hypothesize that, particularly for temporally-intensive datasets, the integration of stacked cross- and self-attention layers provides the necessary capacity to capture and reason about complex temporal dependencies across video frames.

\textit{Takeaway 1: For all subsequent experiments, we use 12-layer pretrained Q-Former w/ SA as our vision-language interface for video understanding with large vision-language model (LVLM).}

\subsection*{Step 2: Temporal-Oriented Training Schemes}

Existing methods \citep{zhang2023video, li2024llama, lin2023video} typically follow a pipeline of pretraining and instruction-tuning, followed by downstream finetuning. In our work, we investigate whether introducing an additional training stage specifically aimed at enhancing temporal understanding can further improve the video comprehension capabilities of LVLMs. To this end, we explore several temporal-oriented training strategies, which are illustrated in Table \ref{tab:step2_temporal_oriented_training_scheme_examples}.
\begin{itemize}
    \item \textbf{Video Captioning (VC):} VC scheme aims to generate compact content of the video by leveraging the encoded information from the video. This objective resembles the next token prediction scheme to pretrain text-only LLM. To implement this objective, we provide the LVLM with a video input and the prompt ``\textit{what does the video describe?}’’, then train it to generate the groundtruth caption. For training data, we utilize 661K video-text pairs from 10M samples of the VIDAL-10M dataset \citep{zhu2023languagebind}.
    \item \textbf{Moment Captioning (MC):} Slightly different from VC, MC aims to caption only a specified part of the video. To implement this objective, we leverage the 745K samples from the InternVid dataset \citep{wang2023internvid}, each of which consists of a query and the specific starting and ending timestamps of the related moment in the video. Based on these timestamps, we convert them to discrete frame indices, then provide the model with the prompt ``\textit{Explain what happened from frame <start> to frame <end> in the video.}''
    \item \textbf{Moment Grounding (MG):} The MG task is the reverse variant of MC. Instead of training the model to write a caption, we let it generate the indices of the start and end frame index of the moment caption. Analogous to MC, we also employ the 745K samples from the InternVid dataset \citep{wang2023internvid}.
    \item \textbf{Dense Captioning (DC):} This task is the more complete and fine-grained version of MC and VC, respectively. In particular, we ask the LVLM ``\textit{Can you give me a breakdown of the occurrences at different timestamps in the video?}''. As Table \ref{tab:step2_temporal_oriented_training_scheme_examples} shows, the model is expected to describe a list of moments with the respective frame indices related to the moment. 
\end{itemize}

{\renewcommand{\arraystretch}{1.2}
\begin{table}[t]
\centering
\caption{Effects of Temporal-Oriented Training Schemes on 7B-LVLM}
\label{tab:step2_temporal_oriented_training_7b}
\resizebox{0.75\linewidth}{!}{
\begin{tabular}{l|c|c|c|c|c|c}
\toprule
\rowcolor{HeaderBlue}
\multicolumn{1}{c|}{\textbf{Training Scheme}} & \textbf{MSRVTT} & \textbf{MSVD} & \textbf{ActivityNet-QA} & \textbf{Breakfast} & \textbf{COIN} & \textbf{LVU} \\ 
\midrule
No temporal training        & 49.4 & 60.8 & 50.6 & 93.1 & 93.4 & 63.8 \\
VC                          & 50.3 & 61.5 & 51.0 & 93.2 & 93.7 & 64.4 \\
MC                          & 51.9 & 62.9 & 52.0 & 93.9 & 93.9 & 64.5 \\
MG                          & 50.9 & 62.3 & 51.4 & 93.3 & 93.9 & 64.6 \\
DC                          & 53.1 & 64.3 & 51.2 & 93.5 & 93.6 & 64.1 \\
\rowcolor{BestRow}
VC + MC + MG + DC           & \textbf{54.5} & \textbf{66.4} & \textbf{52.4} & \textbf{93.7} & \textbf{94.1} & \textbf{65.5} \\ 
\bottomrule
\end{tabular}}
\vspace{-10pt}
\end{table}}

{\renewcommand{\arraystretch}{1.2}
\begin{table}[t]
\centering
\caption{Effects of Temporal-Oriented Training Schemes on 13B-LVLM}
\label{tab:step2_temporal_oriented_training_13b}
\resizebox{0.75\linewidth}{!}{
\begin{tabular}{l|c|c|c|c|c|c}
\toprule
\rowcolor{HeaderBlue}
\multicolumn{1}{c|}{\textbf{Training Scheme}} & \textbf{MSRVTT} & \textbf{MSVD} & \textbf{ActivityNet-QA} & \textbf{Breakfast} & \textbf{COIN} & \textbf{LVU} \\ 
\midrule
No temporal training        & 60.6 & 74.3 & 52.5 & 93.7 & 93.2 & 71.8 \\
VC                          & 62.0 & 74.8 & 53.9 & 94.0 & 93.6 & 72.5 \\
MC                          & 61.2 & 74.6 & 53.0 & 93.5 & 93.9 & 71.9 \\
MG                          & 61.8 & 74.6 & 53.4 & 93.6 & 93.5 & 72.1 \\
DC                          & 62.2 & 75.3 & 54.3 & 94.7 & 94.0 & 72.7 \\
\rowcolor{BestRow}
VC + MC + MG + DC           & \textbf{62.7} & \textbf{75.8} & \textbf{54.5} & \textbf{95.1} & \textbf{94.7} & \textbf{72.8} \\ 
\bottomrule
\end{tabular}}
\vspace{-10pt}
\end{table}}

{\renewcommand{\arraystretch}{1.1}
\begin{table}[t]
\caption{Examples of temporal-oriented training schemes.}
\label{tab:step2_temporal_oriented_training_scheme_examples}
\centering
\resizebox{0.95\linewidth}{!}{
\begin{tabular}{c|c|p{0.4\linewidth}|p{0.4\linewidth}}
\hline
\rowcolor{HeaderBlue}
\textbf{Video} & \multicolumn{1}{c|}{\textbf{Training Scheme}} & \multicolumn{1}{c|}{\textbf{Example prompt}}                                                         & \multicolumn{1}{c}{\textbf{Sample Expected Output}}   \\ \hline
\multirow{2}{*}
   {\includegraphics[width=0.07\linewidth]{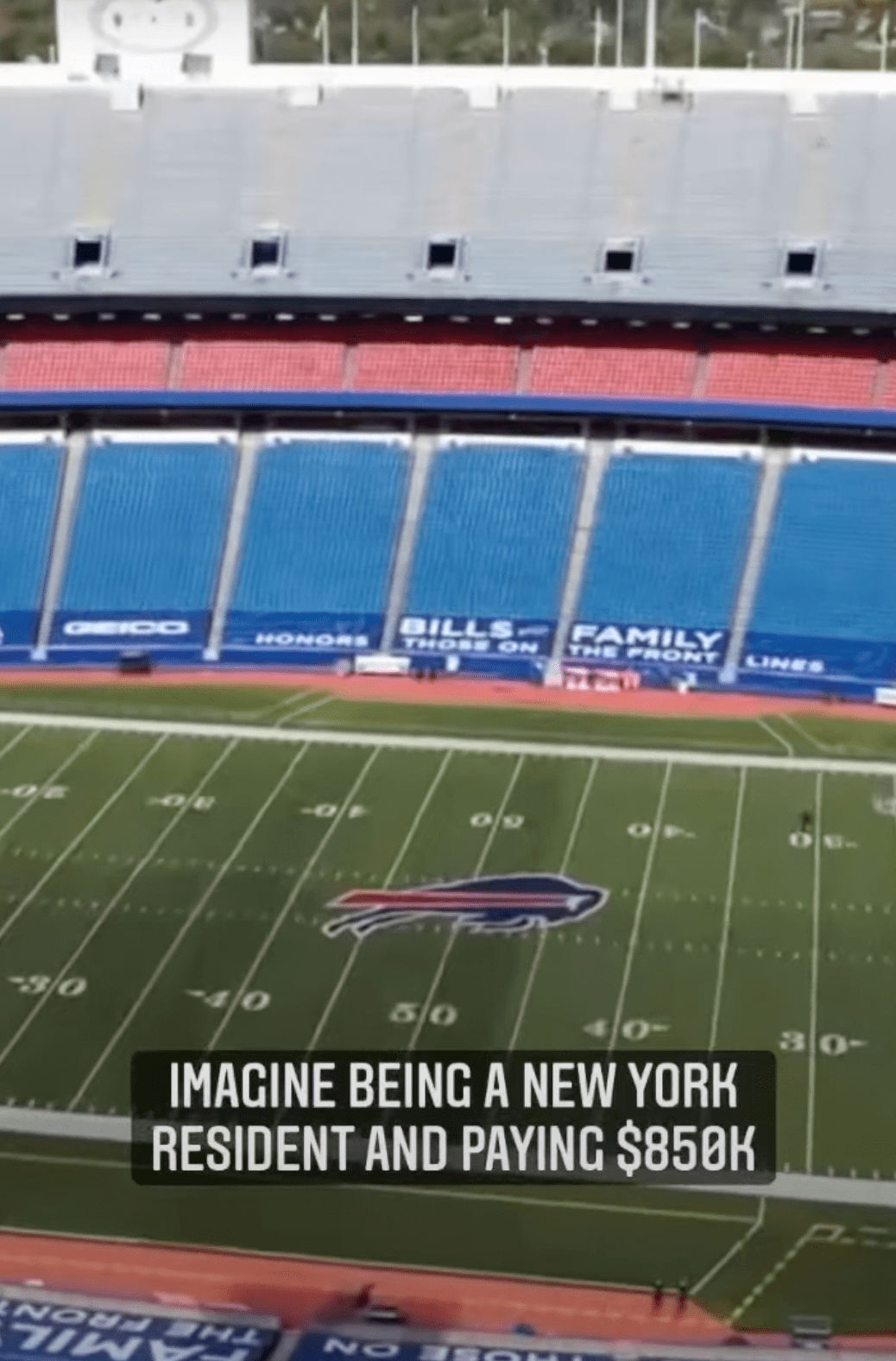} \includegraphics[width=0.07\linewidth]{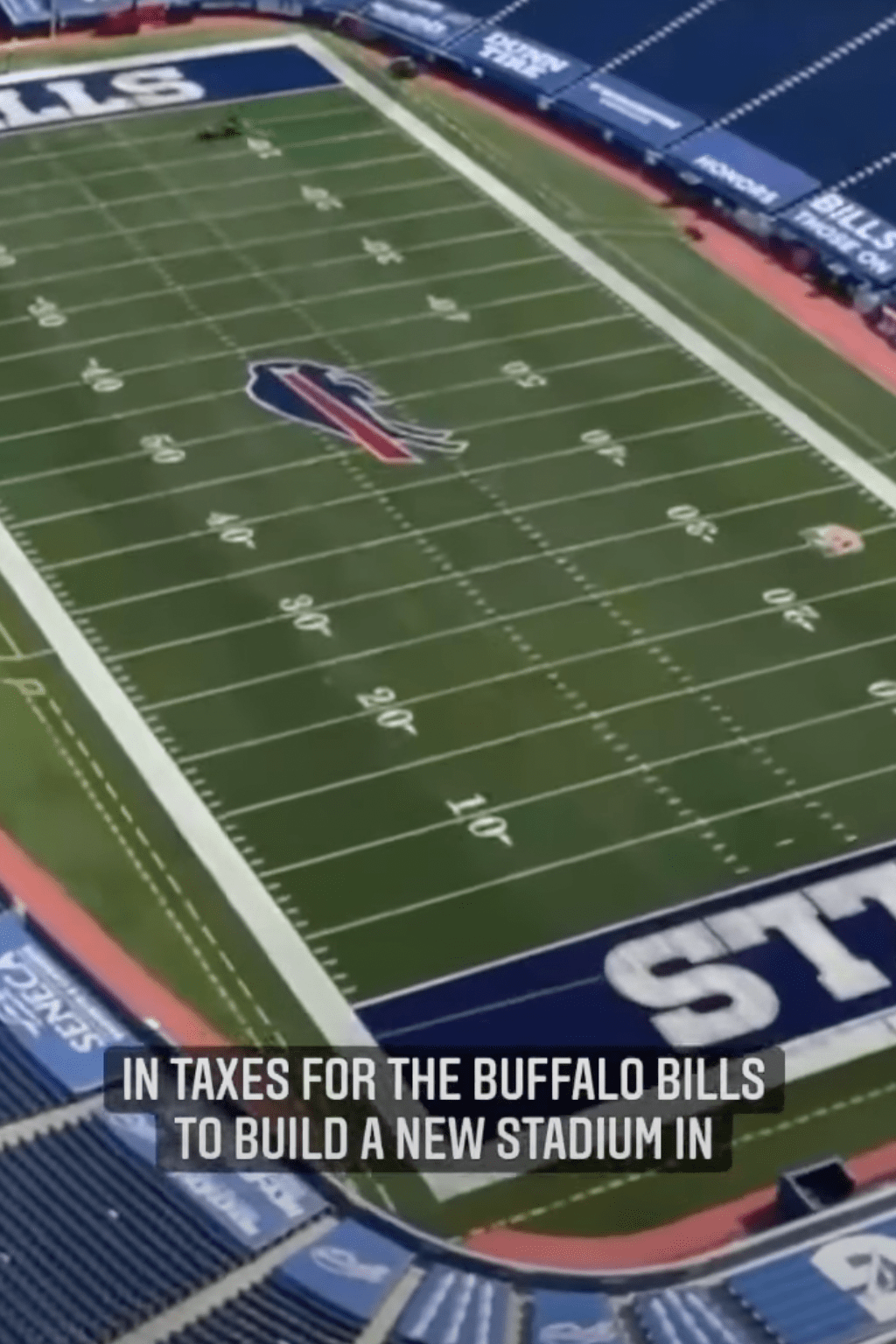}
\includegraphics[width=0.07\linewidth]{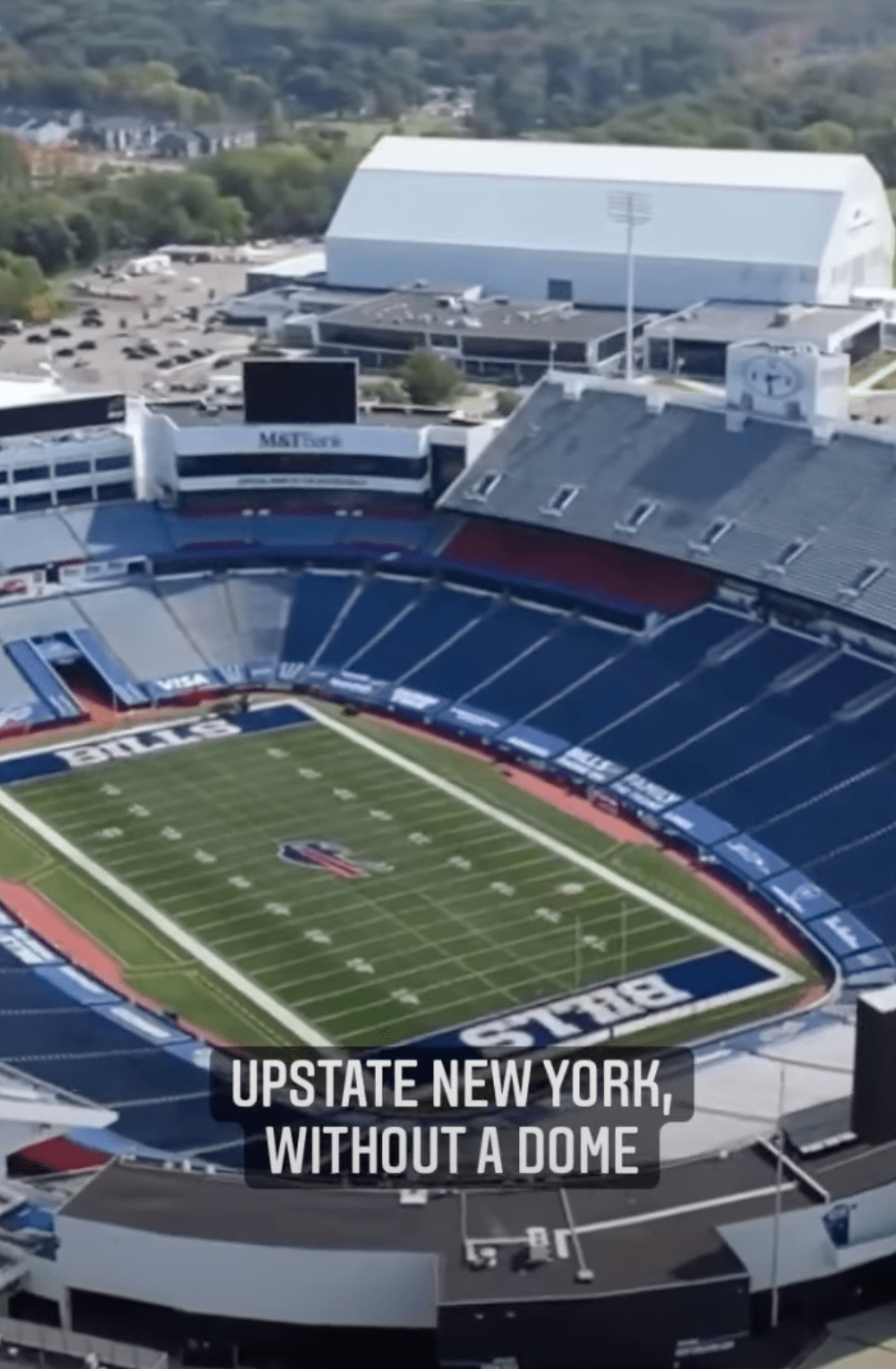} \includegraphics[width=0.07\linewidth]{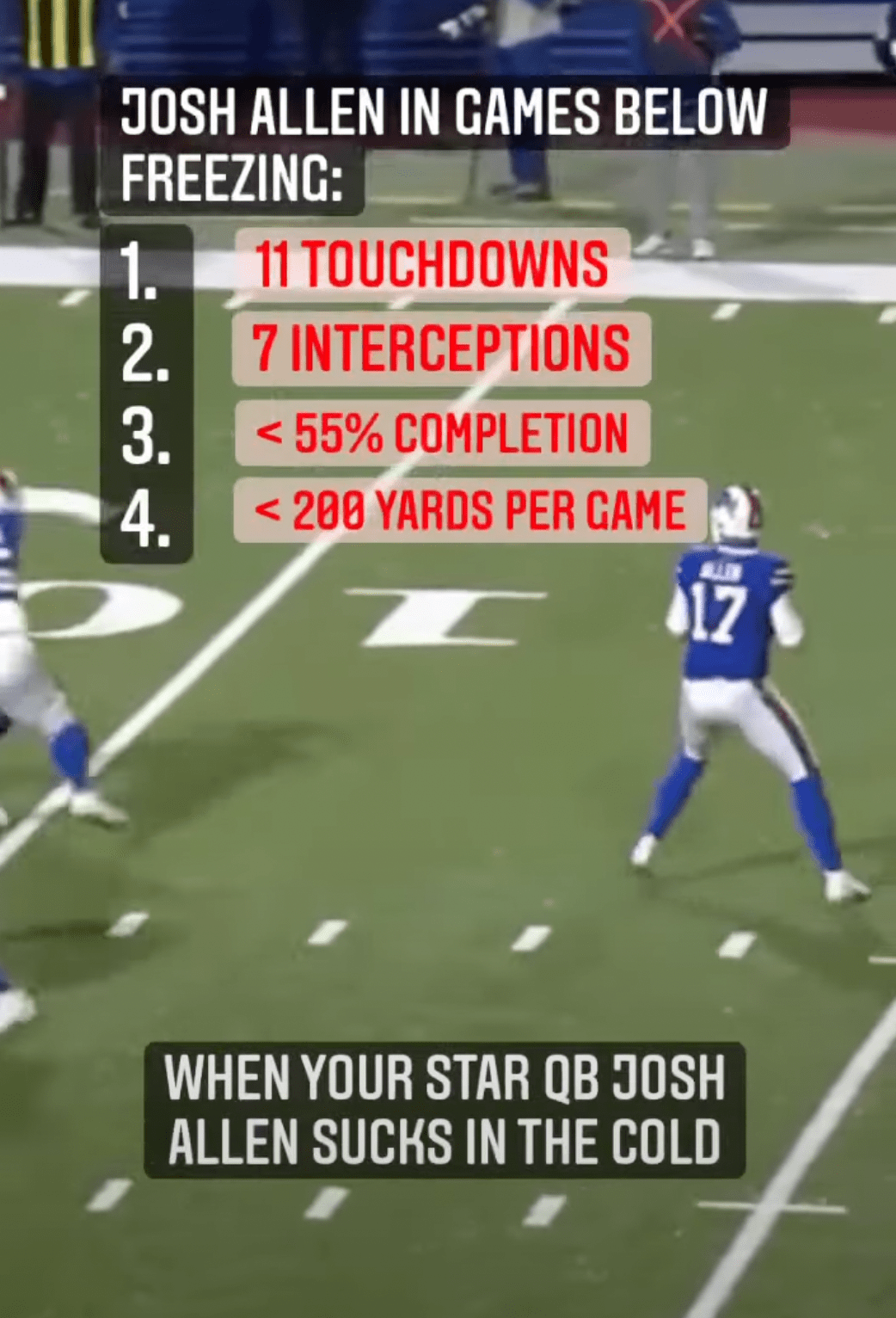}}              & VC                                           & What does the video describe?                                                                             & the buffalo bills new stadium is discussed as being deemed ineffective and not worth the investments made by the city and state.     \\ \hline
   \multirow{5}{*}
   {\includegraphics[width=0.07\linewidth]{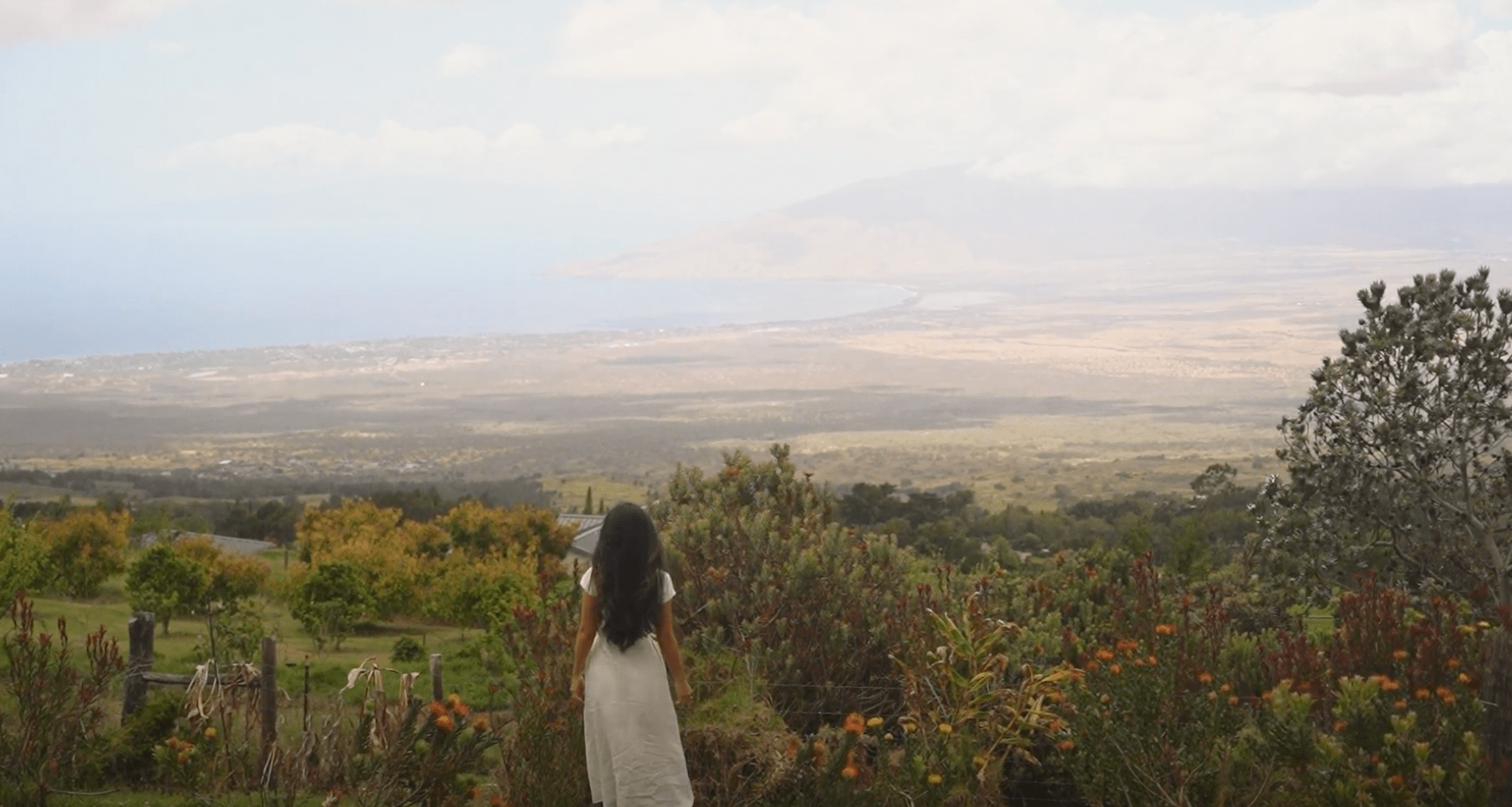} \includegraphics[width=0.07\linewidth]{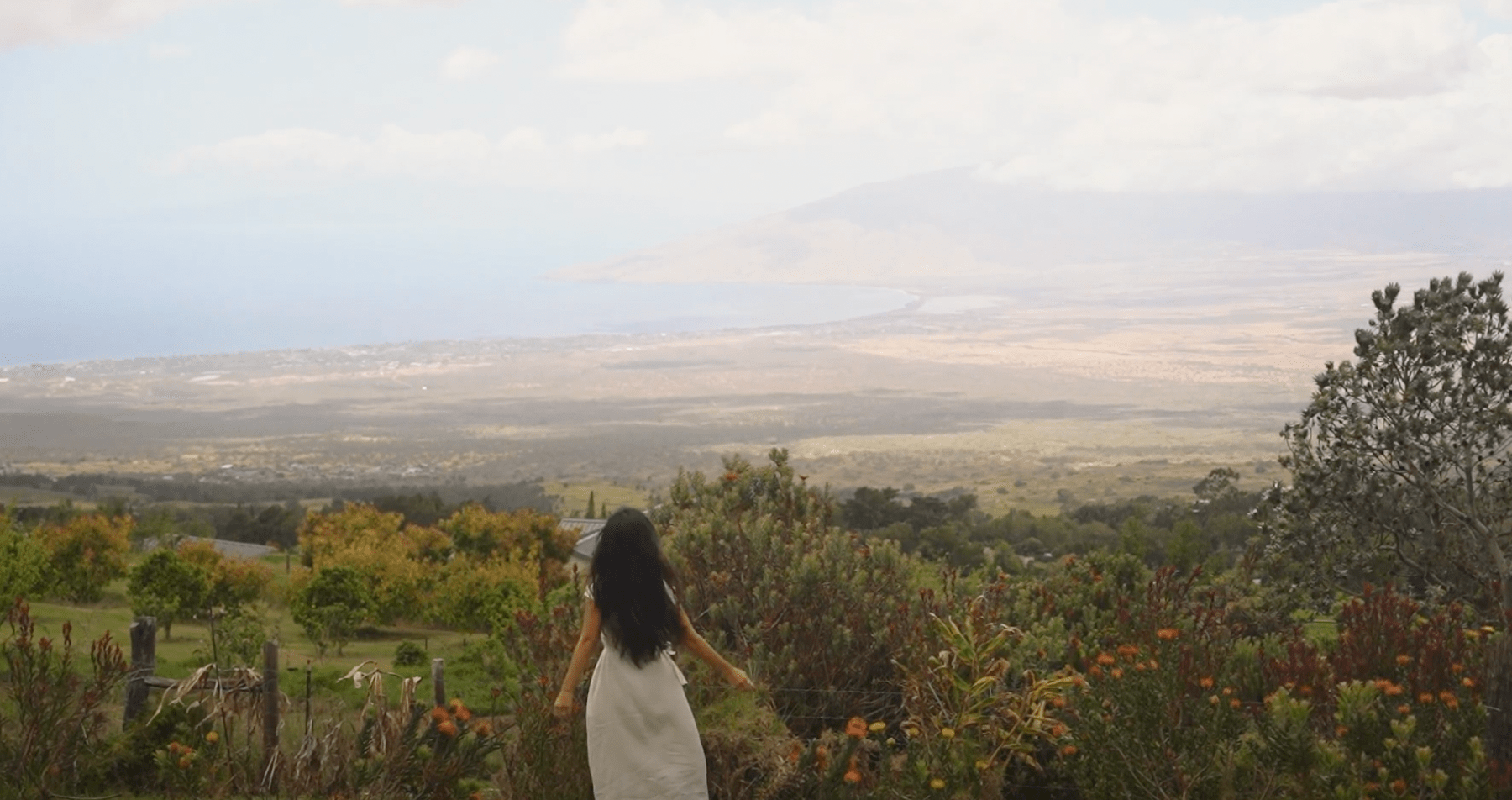}
\includegraphics[width=0.07\linewidth]{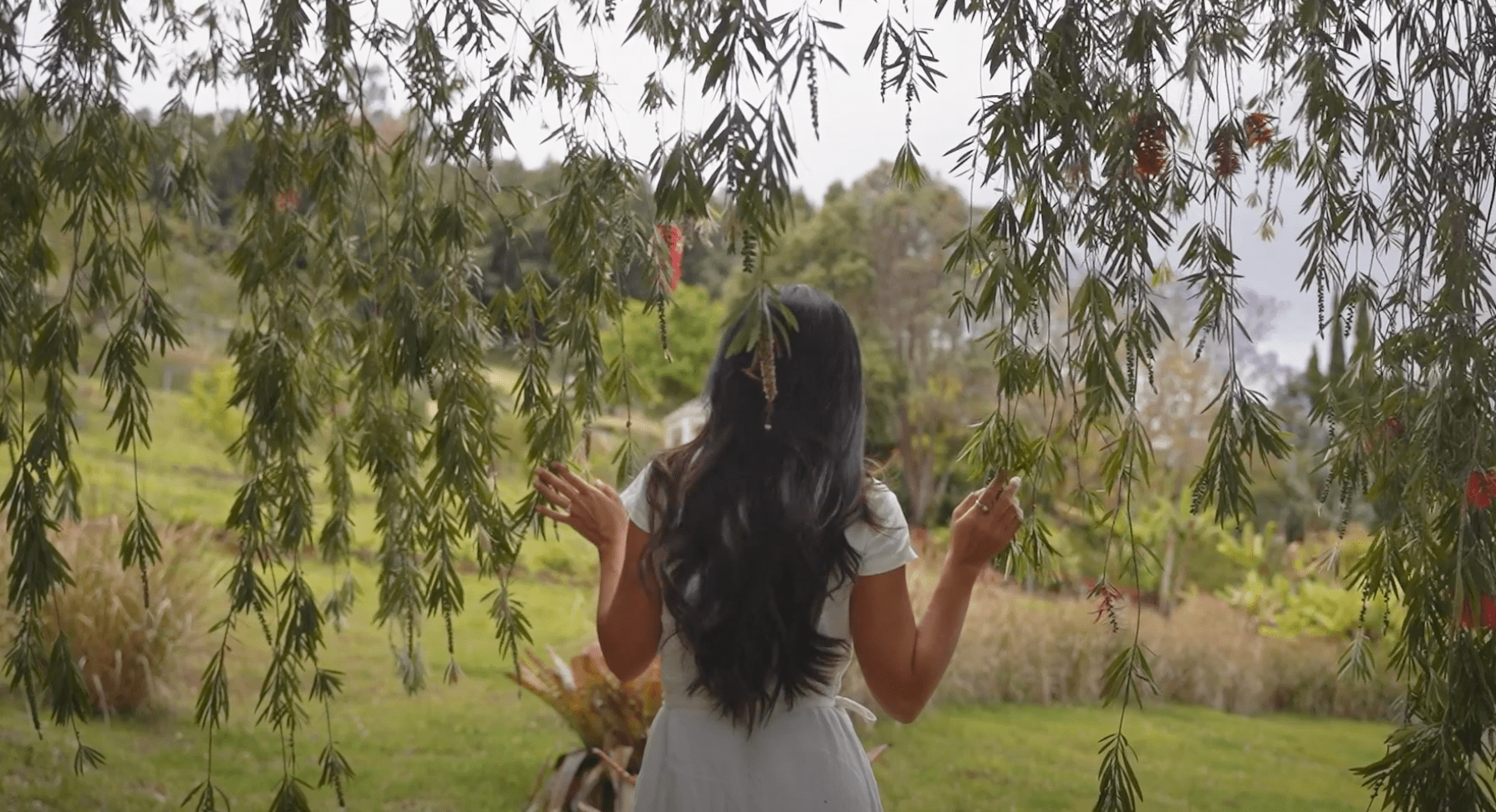} \includegraphics[width=0.07\linewidth]{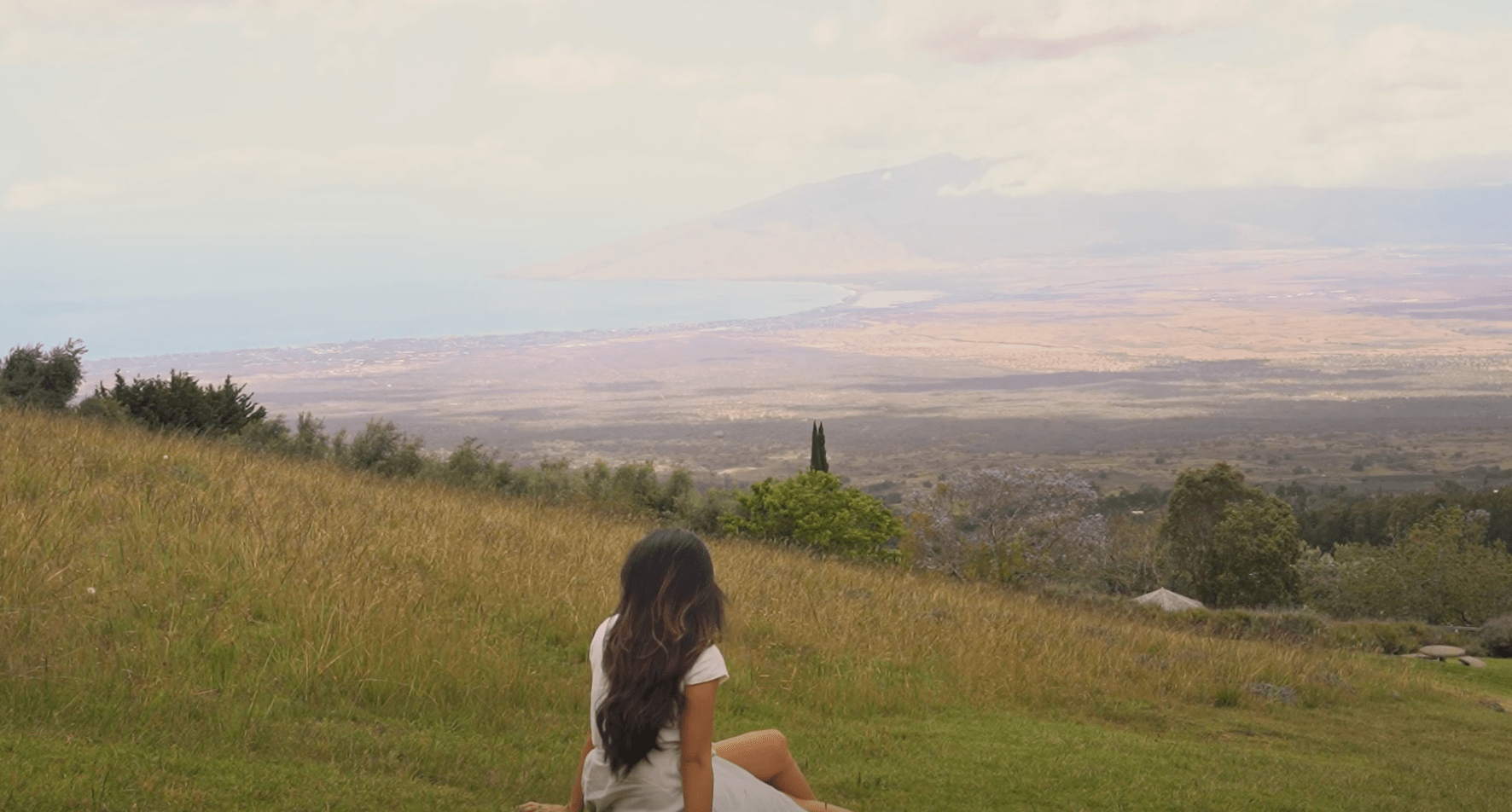}}
  & MC                                           & Explain what happened from frame 1 to frame 4 in the video.                                               & woman in long white dress walking up a hillside path.    \\ \cline{2-4}
 \multirow{5}{*} {\includegraphics[width=0.07\linewidth]{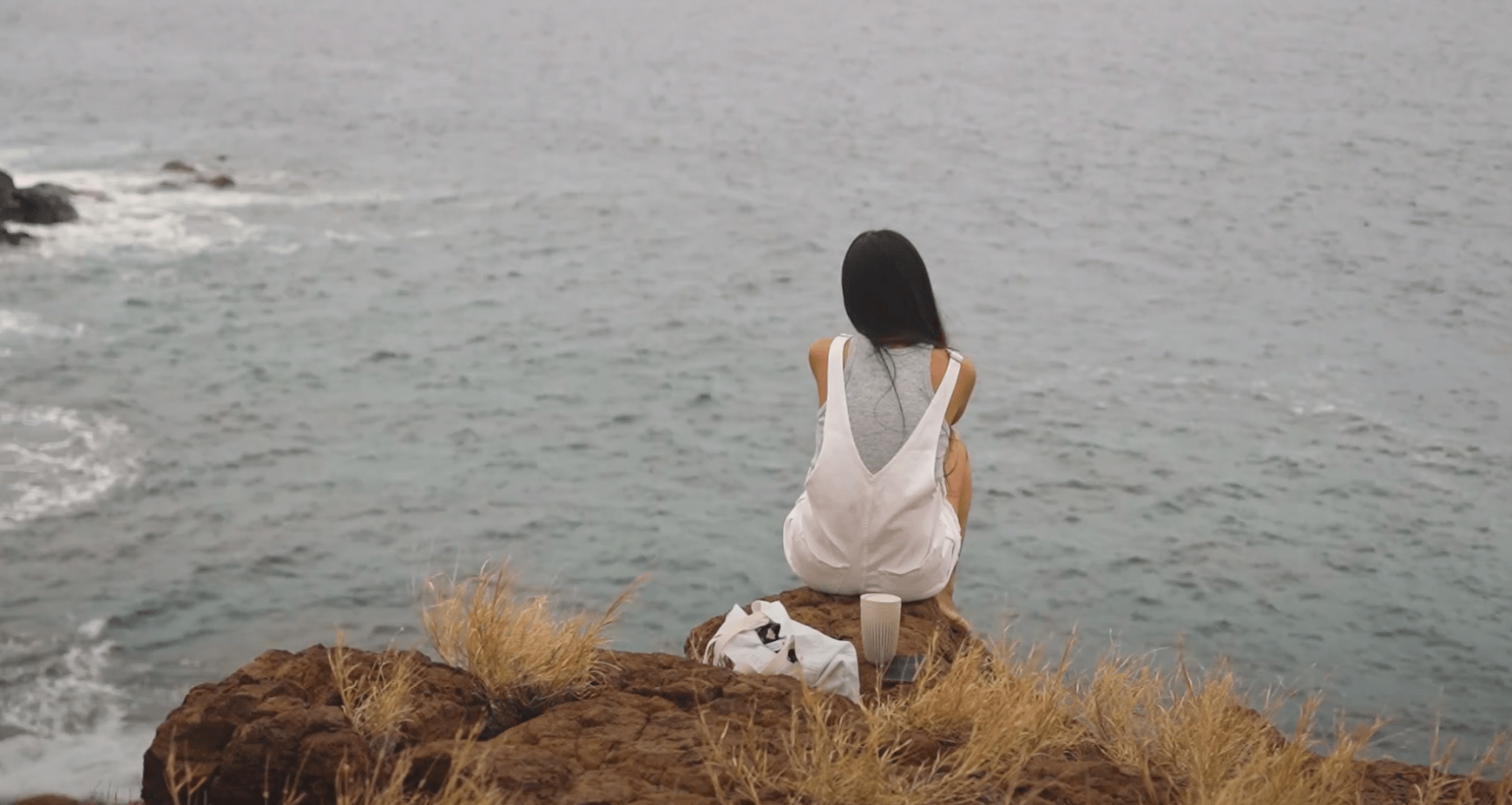} \includegraphics[width=0.07\linewidth]{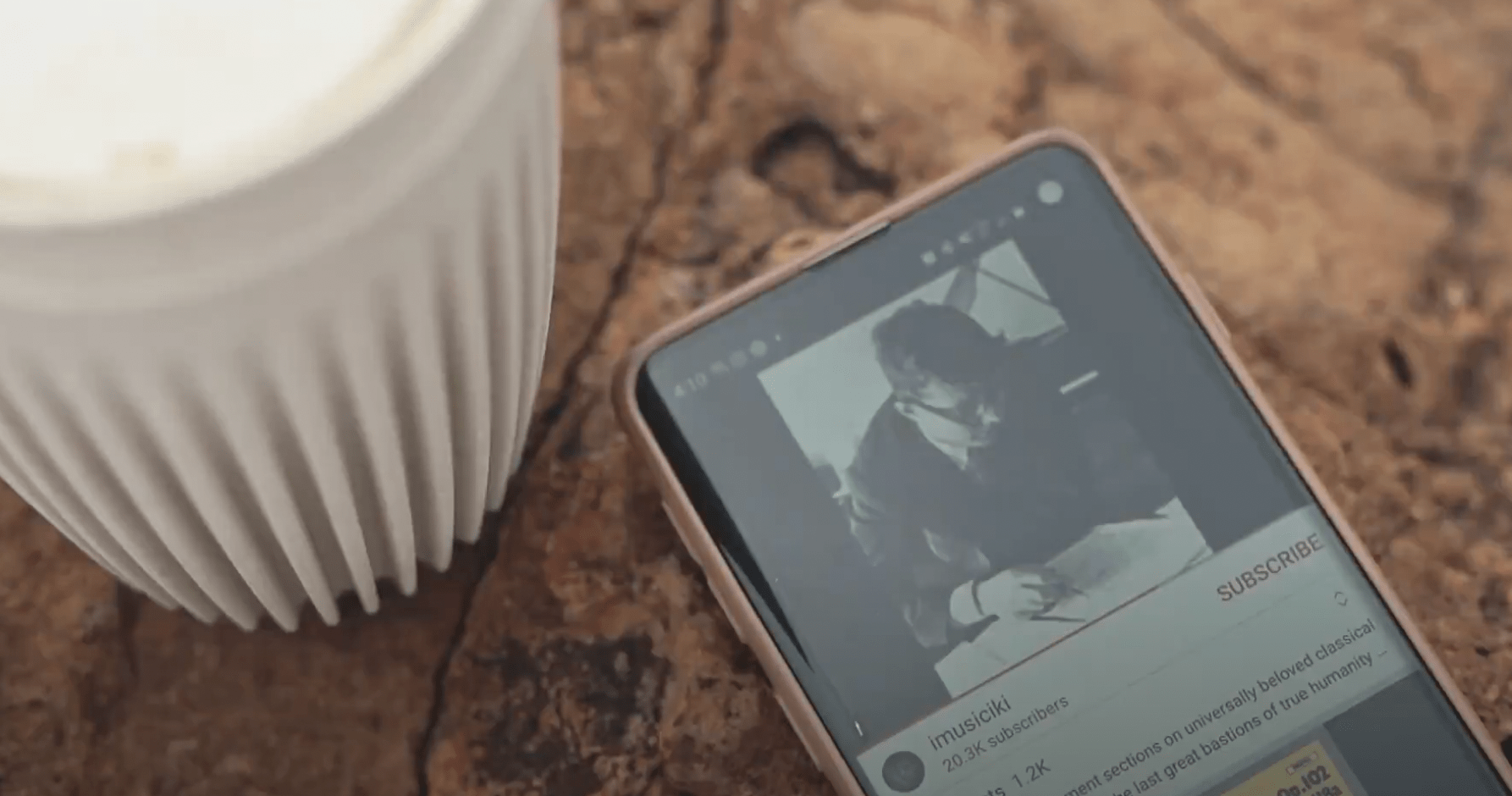}
\includegraphics[width=0.07\linewidth]{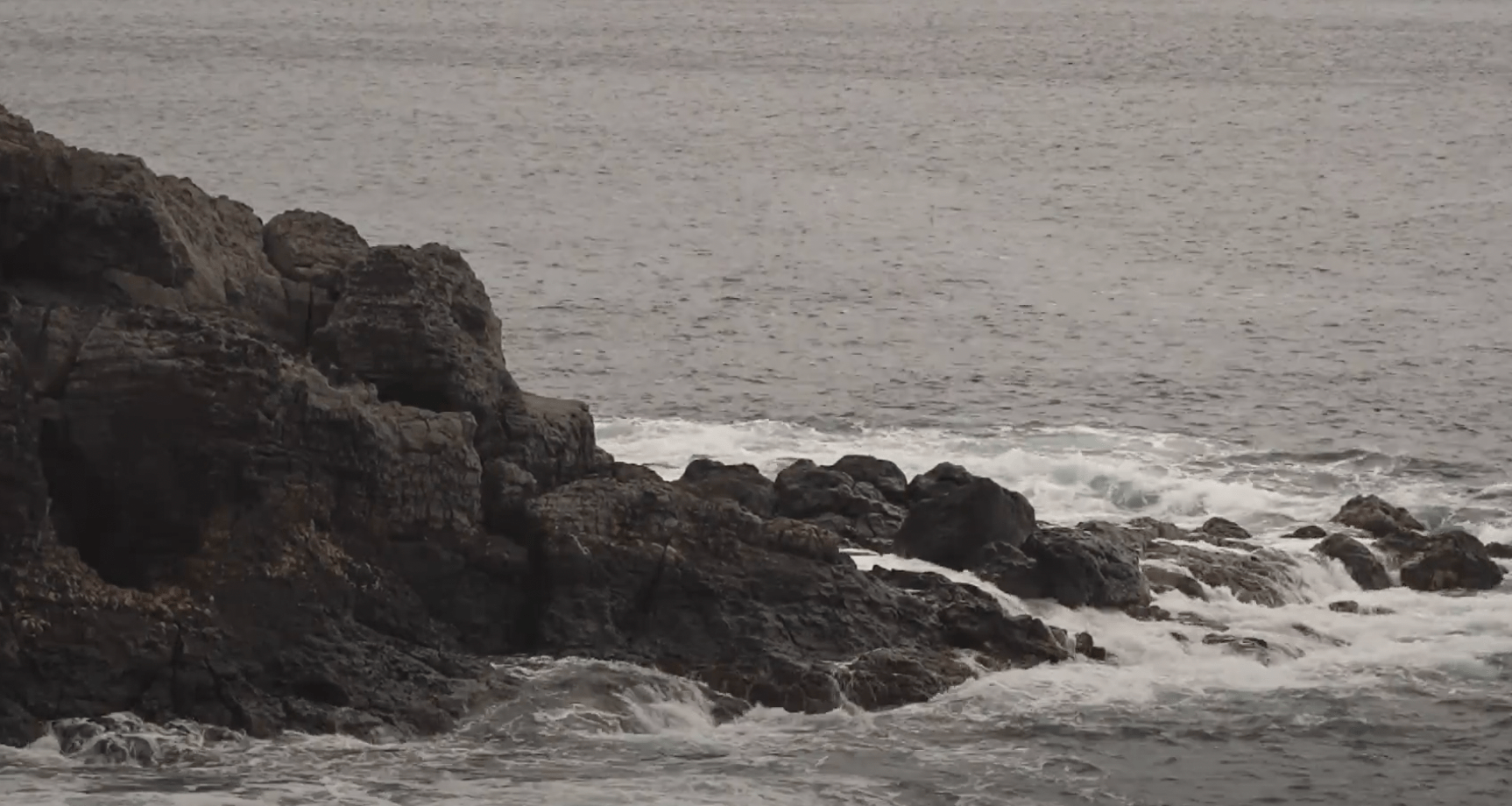} \includegraphics[width=0.07\linewidth]{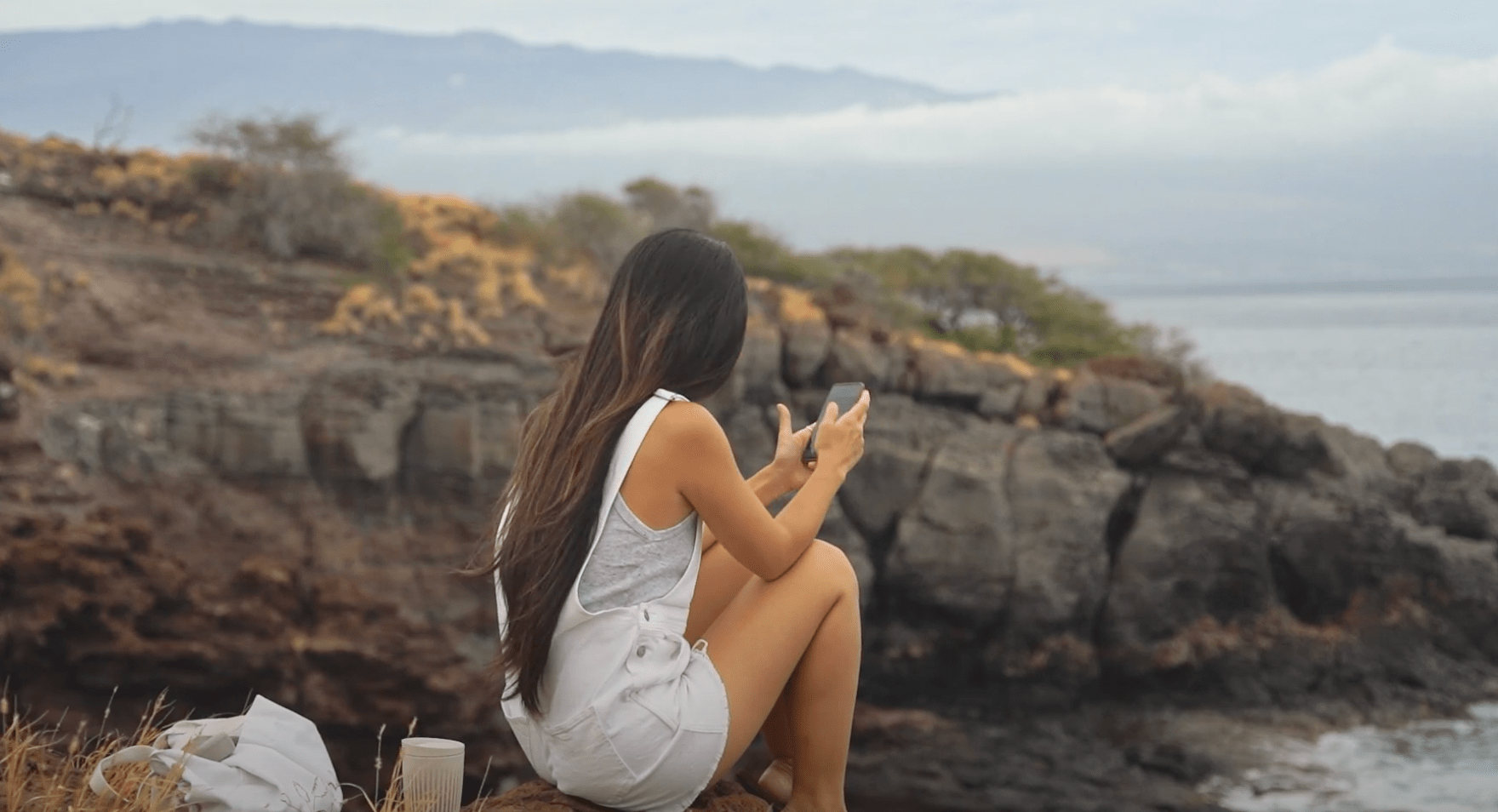}}              & MG                                           & During which frames in the video can we observe ''\textit{woman in long white dress walking up a hillside path}``? & from frame 1 to frame 4                   \\ \cline{2-4}
     \multirow{3}{*} {\includegraphics[width=0.07\linewidth]{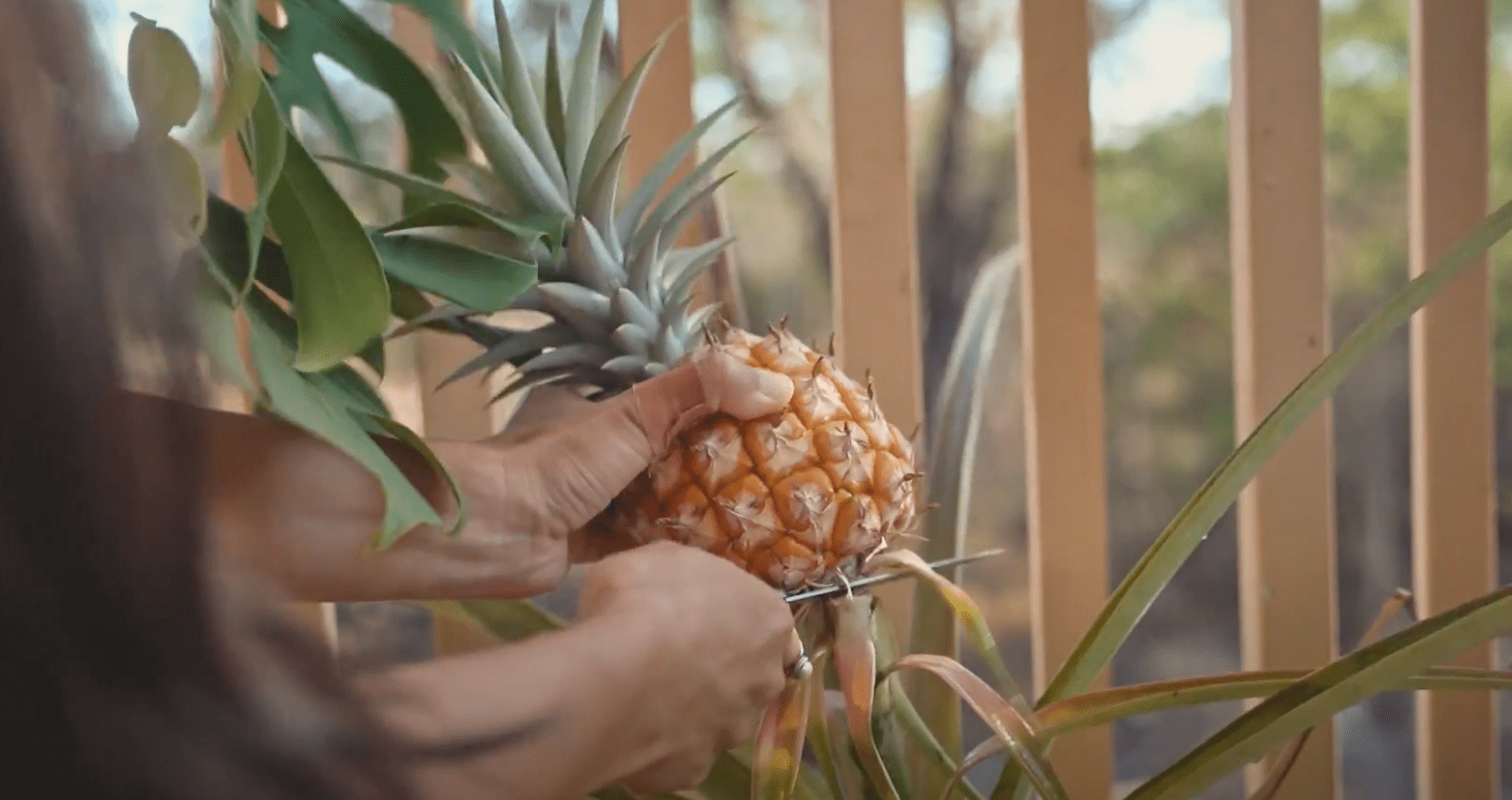} \includegraphics[width=0.07\linewidth]{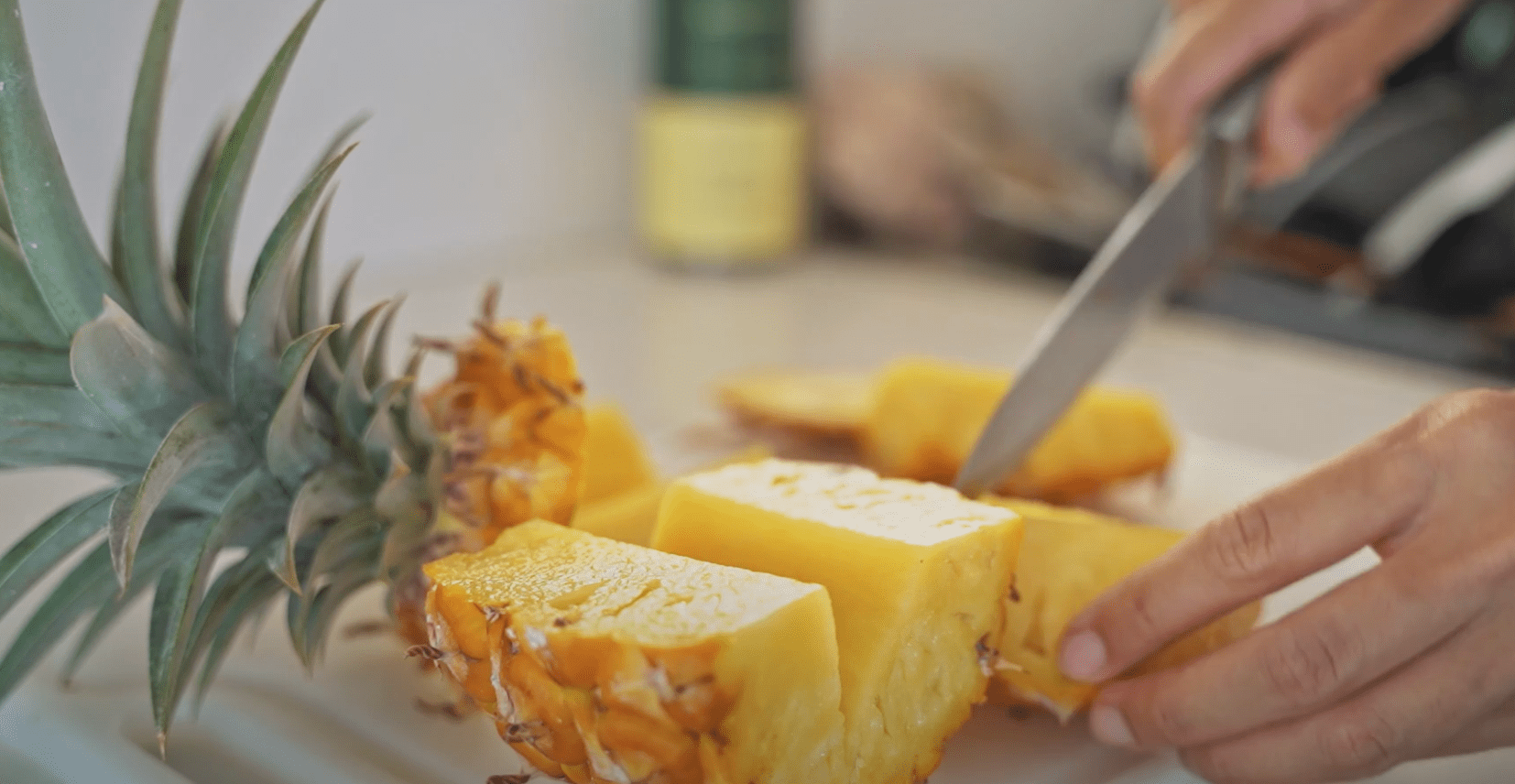}
\includegraphics[width=0.07\linewidth]{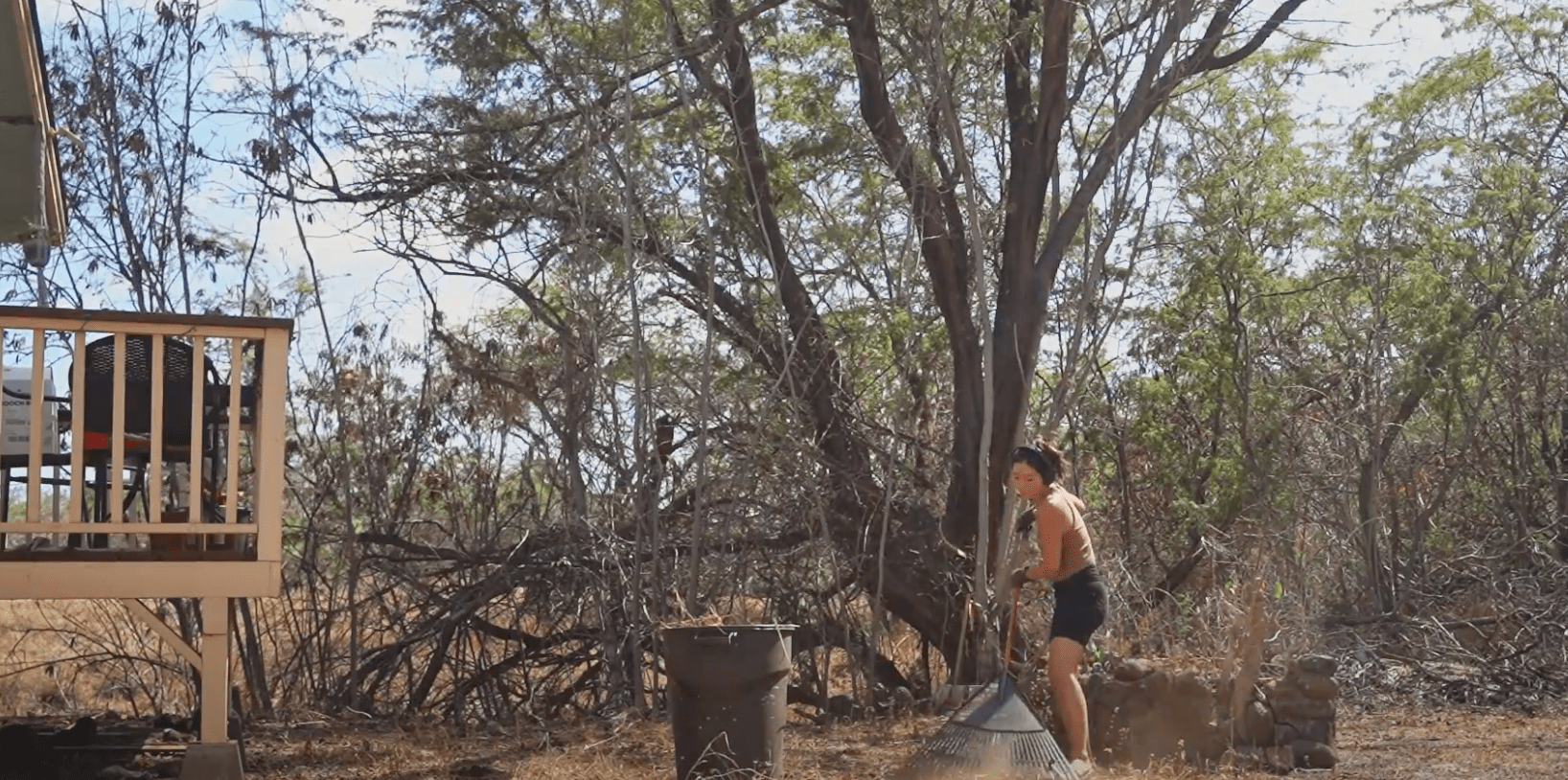} \includegraphics[width=0.07\linewidth]{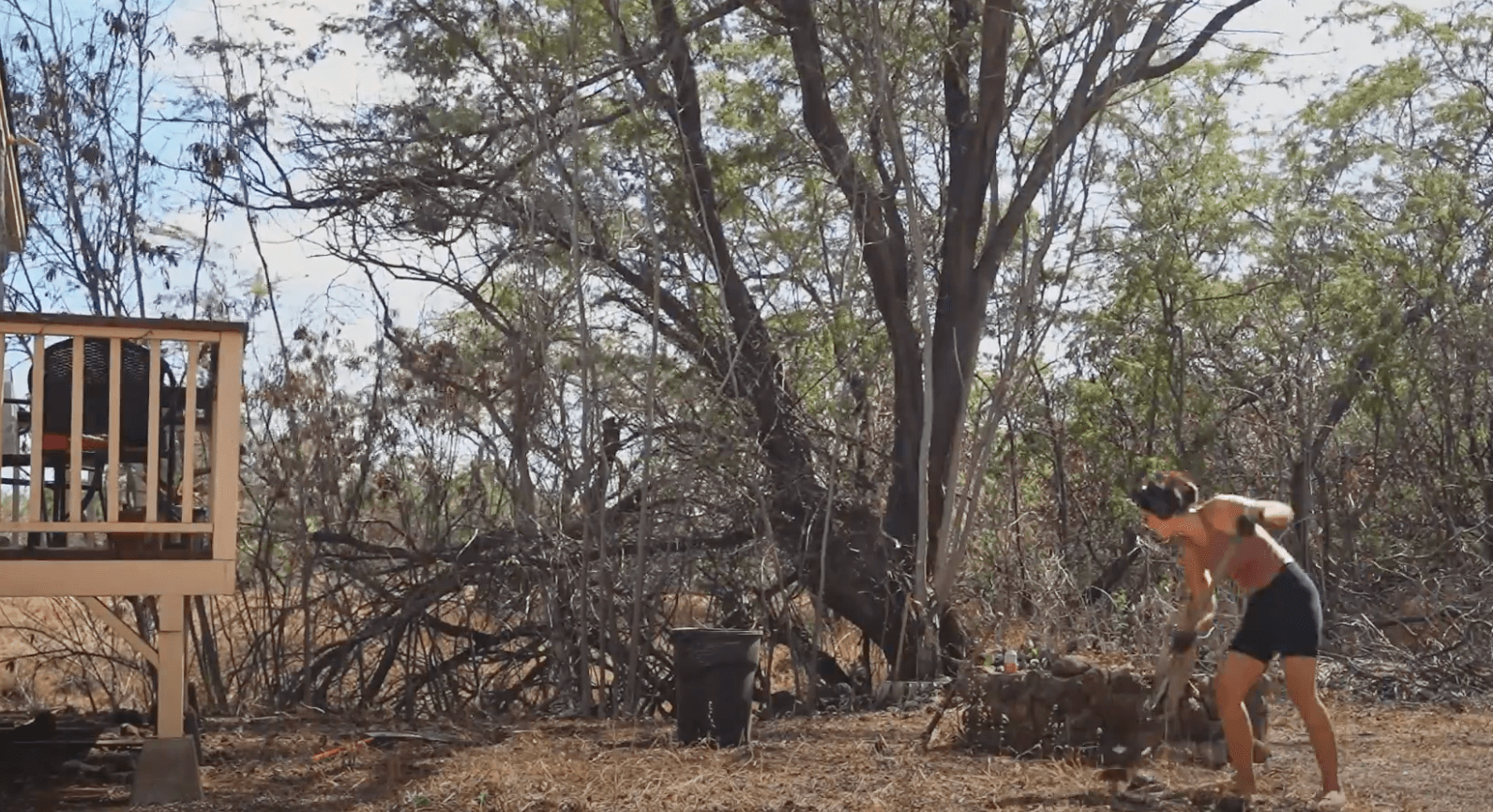}}             & DC                                           & Can you give me a breakdown of the occurrences at different timestamps in the video?                     & woman in long white dress walking up a hillside path, from 1 to 4. a woman sitting on the beach with long hair, from 5 to 8. [...] \\ \cline{2-4}
\hline
\end{tabular}}
\vspace{-15pt}
\end{table}}

Based on the results presented in Table \ref{tab:step2_temporal_oriented_training_7b} and \ref{tab:step2_temporal_oriented_training_13b}, we observe that all temporal-oriented training schemes enhance the temporal understanding capabilities of LVLMs. Notably, the 13B-LVLM shows a more pronounced improvement, particularly when trained with the aggregated scheme VC+MC+MG+DC. This indicates significant untapped potential for even larger LVLMs, especially those exceeeding the 20B parameter scale. Due to computational constraints, we leave the exploration of such large-scale models to future work. 

\textit{Takeaway 2: For the remaining experiments, we add an additional temporal-oriented training stage and use VC, MC, MG, and DC as training schemes.}

\vspace{-5pt}
\subsection*{Step 3: Memory Bank for Video Representations}
{\renewcommand{\arraystretch}{1.2}
\begin{table}[t]
\centering
\caption{Effect of Memory Bank on 7B-LVLM}
\label{tab:step3_memory_bank_7b}
\resizebox{0.75\linewidth}{!}{
\begin{tabular}{l|c|c|c|c|c|c}
\toprule
\rowcolor{HeaderBlue}
\textbf{Memory Bank ($B$)} & \textbf{MSRVTT} & \textbf{MSVD} & \textbf{ActivityNet-QA} & \textbf{Breakfast} & \textbf{COIN} & \textbf{LVU} \\ 
\midrule
$B = 0$  & 54.5 & 66.4 & 51.4 & 93.7 & 94.1 & 65.5 \\
$B = 10$ & 56.2 & 68.2 & 51.9 & 93.9 & 94.1 & 66.0 \\
$B = 20$ & 60.7 & 72.5 & 52.5 & 94.3 & 94.7 & 68.1 \\
$B = 30$ & 60.7 & 72.5 & 52.5 & 94.7 & 94.6 & 68.1 \\
$B = 40$ & 60.7 & 72.6 & 52.5 & 94.9 & 94.8 & 68.2 \\
$B = 50$ & 60.8 & 72.6 & 52.6 & 94.9 & 94.9 & 68.2 \\
\rowcolor{BestRow}
$B = 60$ & \textbf{60.8} & \textbf{72.6} & \textbf{52.6} & \textbf{95.0} & \textbf{95.0} & \textbf{68.3} \\ 
\bottomrule
\end{tabular}}
\vspace{-15pt}
\end{table}}

{\renewcommand{\arraystretch}{1.2}
\begin{table}[t]
\centering
\caption{Effect of Memory Bank on 13B-LVLM}
\label{tab:step3_memory_bank_13b}
\resizebox{0.75\linewidth}{!}{
\begin{tabular}{l|c|c|c|c|c|c}
\toprule
\rowcolor{HeaderBlue}
\textbf{Memory Bank ($B$)} & \textbf{MSRVTT} & \textbf{MSVD} & \textbf{ActivityNet-QA} & \textbf{Breakfast} & \textbf{COIN} & \textbf{LVU} \\ 
\midrule
$B = 0$  & 62.7 & 75.8 & 54.5 & 95.1 & 94.7 & 72.8 \\
$B = 10$ & 63.3 & 75.9 & 54.9 & 95.3 & 94.9 & 73.1 \\
$B = 20$ & 63.5 & 76.3 & 55.2 & 96.0 & 95.0 & 73.4 \\
$B = 30$ & 63.5 & 76.3 & 55.4 & 96.5 & 95.6 & 73.9 \\
\rowcolor{BestRow}
$B = 40$ & \textbf{64.4} & \textbf{77.5} & \textbf{55.8} & \textbf{97.6} & \textbf{96.8} & \textbf{74.7} \\
$B = 50$ & 63.9 & 77.0 & 55.6 & 96.8 & 96.0 & 74.6 \\
$B = 60$ & 63.5 & 76.5 & 55.5 & 96.7 & 96.0 & 74.3 \\ 
\bottomrule
\end{tabular}}
\vspace{-15pt}
\end{table}}

Building upon the model developed in Step 2, we further investigate how the LVLM processes video inputs. A straightforward approach involves encoding visual frames or patches and concatenating their representations along the temporal axis. However, the limited context length limit of the LVLM, coupled with GPU memory constraints, restricts the number of video frames that can be processed simultaneously. An alternative strategy is to apply temporal pooling \citep{maaz2023video, xu2024pllava}, but as demonstrated in our Step 1 analysis, this leads to suboptimal performance. Instead, we propose a different approach, \textit{i.e.} processing video frames sequentially and storing their features in a memory bank. We conduct an ablation study on the size of the memory bank $B \in \{10, 20, 30, 40, 50, 60\}$ and present our findings in Table \ref{tab:step3_memory_bank_7b} and \ref{tab:step3_memory_bank_13b}.

Based on these results, we find that incorporating a memory bank is an effective strategy, consistently outperforming standard pooling methods. Randomly sampling a fixed number of frames also proves suboptimal, particularly for long-term temporal understanding, as the sampled frames might fail to capture critical video context. Lastly, we note that increasing the memory bank size yields more significant improvement for the 13B-LVLM than the 7B-LVLM. This indicates that larger-scale models possess greater capacity to absorb and utilize richer video information. 

\textit{Takeaway 3: For our remaining experiments, we add a memory bank for video encoding.}

\vspace{-5pt}
\subsection*{Step 4: Mixture-of-Experts for Q-Former}
Building upon Step 3, we next explore strategies to enhance the capacity of the vision-language interface, which plays a critical role in conveying video information to the LLM. Given that naively adding randomly initialized layers tends to yield suboptimal performance---as demonstrated in Step 1---we turn to the mixture-of-experts (MoE) approach. An MoE module consists of a router and a set of experts, where each expert is a feedforward network. The router typically comprises a linear projection followed by a gating function, \textit{e.g.} ReLU or Softmax, to compute the probabilities for routing a query token to specific experts. When a token encounters the MoE, the router selects a subset of experts to process the token, and their outputs are combined additively. This technique allows us to expand the parameter capacity of the Q-Former while keeping computational cost and latency manageable, as the model activates only a fraction of the total parameters for each token.

The exploration of MoE has remained scarce for LVLMs, especially for the vision-language interface, even though it has been investigated extensively in LLMs \citep{cai2024survey}. In our work, we will experiment with the following categories of MoE:
\vspace{-5pt}
\begin{itemize}
    \itemsep -1pt
    \item \textbf{Dense MoE:} the dense MoE activates all expert networks during each iteration. Based on the probability that the router produces for each expert, the outputs for an input token will be aggregated accordingly. 

    \item \textbf{Sparse MoE:} to reduce computational overhead, we can activate only a subset of experts during each forward pass. To achieve this sparsity, we can compute a weighted sum of the expert outputs from only the top-$k$ experts, rather than combining the outputs from all experts. In our work, we experiment with top-$k$ where $k = 1$ or $k = 2$.
\end{itemize}

For each type of MoE, we ablate the number of experts $E \in \{2, 4, 8\}$. In addition to Q-Former, we also add MoE to LLM to comprehensively study its effect on the LVLM.

{\renewcommand{\arraystretch}{1.2}
\begin{table}[h!]
\centering
\caption{Effect of Mixture-of-Experts (MoE) on 7B-LVLM}
\label{tab:step4_moe_lvlm_7b}
\resizebox{0.9\linewidth}{!}{
\begin{tabular}{llc|ccccccc}
\toprule
\rowcolor{HeaderBlue}
\textbf{Position of MoE} & \multicolumn{1}{c}{\textbf{Type}} & \textbf{Experts ($E$)} & \textbf{MSRVTT} & \textbf{MSVD} & \textbf{ActivityNet-QA} & \textbf{Breakfast} & \textbf{COIN} & \textbf{LVU} \\ 
\midrule
Q-Former & Sparse & 2 & 63.9 & 76.9 & 56.3 & 95.6 & 95.4 & 72.9 \\
         &        & 4 & 64.1 & 77.6 & 56.7 & 96.3 & 96.1 & 73.7 \\
         &        & 8 & \textbf{65.0} & \textbf{78.1} & \textbf{57.3} & \textbf{97.0} & \textbf{96.6} & \textbf{74.5} \\
\rowcolor{GrayRow}
         & Dense  & 2 & 63.9 & 76.4 & 55.7 & 95.2 & 94.9 & 72.3 \\
\rowcolor{GrayRow}
         &        & 4 & 64.4 & 76.8 & 56.6 & 96.0 & 95.8 & 73.0 \\
\rowcolor{GrayRow}
         &        & 8 & 64.9 & 77.3 & 57.0 & 96.1 & 96.2 & 73.4 \\
\midrule
LLM      & Sparse & 2 & 54.9 & 72.9 & 55.4 & 93.4 & 92.0 & 72.5 \\
         &        & 4 & 54.9 & 73.5 & 55.5 & 93.6 & 92.2 & 72.7 \\
         &        & 8 & 55.4 & 73.6 & 55.9 & 93.6 & 92.5 & 72.8 \\
\rowcolor{GrayRow}
         & Dense  & 2 & 54.5 & 72.4 & 54.9 & 93.1 & 91.7 & 72.4 \\
\rowcolor{GrayRow}
         &        & 4 & 54.8 & 72.7 & 55.2 & 93.4 & 91.8 & 72.6 \\
\rowcolor{GrayRow}
         &        & 8 & 54.9 & 73.1 & 55.6 & 93.7 & 92.0 & 72.9 \\
\bottomrule
\end{tabular}}
\vspace{-5pt}
\end{table}}

{\renewcommand{\arraystretch}{1.2}
\begin{table}[h!]
\centering
\vspace{-5pt}
\caption{Effect of Mixture-of-Experts (MoE) on 13B-LVLM}
\label{tab:step4_moe_lvlm_13b}
\resizebox{0.85\linewidth}{!}{
\begin{tabular}{llc|ccccccc}
\toprule
\rowcolor{HeaderBlue}
\textbf{Position of MoE} & \multicolumn{1}{c}{\textbf{Type}} & \textbf{Experts ($E$)} & \textbf{MSRVTT} & \textbf{MSVD} & \textbf{ActivityNet-QA} & \textbf{Breakfast} & \textbf{COIN} & \textbf{LVU} \\ 
\midrule
Q-Former & Sparse & 2 & 65.2 & 78.4 & 57.2 & 97.6 & 97.5 & 75.5 \\
         &        & 4 & 65.6 & 78.8 & 57.6 & 97.8 & 97.7 & 75.9 \\
         &        & 8 & \textbf{66.7} & \textbf{79.5} & \textbf{58.3} & \textbf{98.5} & \textbf{97.8} & \textbf{76.1} \\
\rowcolor{GrayRow}
         & Dense  & 2 & 65.0 & 77.3 & 56.1 & 96.4 & 96.5 & 74.7 \\
\rowcolor{GrayRow}
         &        & 4 & 65.5 & 77.7 & 56.8 & 96.6 & 96.8 & 74.9 \\
\rowcolor{GrayRow}
         &        & 8 & 66.2 & 78.3 & 57.3 & 97.6 & 97.0 & 75.7 \\
\midrule
LLM      & Sparse & 2 & 59.3 & 73.3 & 53.4 & 91.9 & 92.9 & 69.1 \\
         &        & 4 & 59.8 & 73.6 & 53.7 & 93.1 & 93.6 & 69.2 \\
         &        & 8 & 60.2 & 73.7 & 54.1 & 93.2 & 94.2 & 69.6 \\
\rowcolor{GrayRow}
         & Dense  & 2 & 58.7 & 72.7 & 52.9 & 92.2 & 92.3 & 71.9 \\
\rowcolor{GrayRow}
         &        & 4 & 59.0 & 72.9 & 53.0 & 92.3 & 92.5 & 70.0 \\
\rowcolor{GrayRow}
         &        & 8 & 59.4 & 73.7 & 53.9 & 92.3 & 93.1 & 70.4 \\
\bottomrule
\end{tabular}}
\vspace{-5pt}
\end{table}}

Based on the results in Table \ref{tab:step4_moe_lvlm_7b} and \ref{tab:step4_moe_lvlm_13b}, we observe that integrating MoE into Q-Former leads to a substantial boost in video understanding performance. Moreover, we note that both sparse and dense MoE categories bring improvement, with sparse MoE being slightly more effective. We hypothesize that sparse MoE provides a higher degree of specialization for LVLM to handle specific types of temporal circumstances. Perhaps unsurprisingly, scaling up MoE with more experts puts more significant impact to the 13B-LVLM than the 7B-LVLM, which implies further potential for LVLM in the upscaling direction. On the other hand, adding MoE to LLM degrades the performance. This indicates that MoEs might tamper with the pre-trained knowledge in LLM.

\textit{Final takeway: Our final scaled-up temporal-oriented LVLM improves the initial LVLM baseline by 10.1\% and 5.1\% in terms of the 7B and 13B variant, respectively.}
\vspace{-5pt}
\section{Experimental Results}
\vspace{-5pt}
\label{sect:experimental_results}
We validate our temporal-oriented recipe on two popular video understanding tasks, \textit{i.e.} video question answering and video captioning. All of our experiments are conducted using 8 H100 GPUs. For implementation details and dataset descriptions, we refer our readers to Appendix \ref{app:implementation_details} and Appendix \ref{app:dataset_descriptions}, respectively. 

\noindent\textbf{Video question answering.} We compare our results with existing methods on six datasets MSRVTT \citep{xu2016msr}, MSVD \citep{chen2011collecting}, ActivityNet-QA \citep{krishna2017dense}, Breakfast \citep{kuehne2014language}, COIN \citep{tang2019coin}, and LVU \citep{wu2021towards} in Table \ref{tab:experimental_results}. Our method substantially outperforms previous approaches on multiple datasets, achieving average accuracies of 66.7\% (\textbf{+18.2\%}), 79.5\% (\textbf{+18.9\%}), 58.3\% (\textbf{+8.5\%}), 98.5\% (\textbf{+5.5\%}), 97.8\% (\textbf{+4.6\%}), and 76.1\% (\textbf{+13.1\%}) on MSRVTT, MSVD, ActivityNet-QA, Breakfast, COIN, and LVU, respectively. 

\noindent\textbf{Video captioning.} We present our results for the video captioning task on MSRVTT \citep{xu2016msr} and MSVD \citep{chen2011collecting}. Our results indicate that we significantly improve upon previous works by a large margin. Particularly, we outperform existing methods by \textbf{20.8\%} on MSRVTT and \textbf{21.0\%} on MSVD.

{\renewcommand{\arraystretch}{1.2}
\begin{table}[h!]
\vspace{-10pt}
\centering
\caption{Comparison with existing methods on Video Question Answering (VideoQA) and Video Captioning tasks. The best results are in \textbf{bold}, and the second-best are \underline{underlined}.}
\label{tab:experimental_results}
\resizebox{0.85\linewidth}{!}{
\begin{tabular}{l|cccccc|cc}
\toprule
\rowcolor{HeaderBlue}
\multicolumn{1}{c|}{\textbf{Method}}  & \multicolumn{6}{c|}{\textbf{VideoQA}} & \multicolumn{2}{c}{\textbf{Video Captioning}} \\
\rowcolor{HeaderBlue}
& \textbf{MSRVTT} & \textbf{MSVD} & \textbf{ActivityNet-QA} & \textbf{Breakfast} & \textbf{COIN} & \textbf{LVU} & \textbf{MSRVTT} & \textbf{MSVD} \\ 
\midrule
MA-LMM \citep{he2024ma}             & 48.5 & 60.6 & 49.8 & 93.0 & 93.2 & 63.0 & 43.3 & 49.1 \\
VALLEY \citep{luo2023valley}        & 50.8 & 69.2 & 44.9 & 83.8 & 84.0 & 56.8 & 39.0 & 44.3 \\
LLaMA-VID \citep{li2024llama}       & 58.9 & 70.0 & 47.5 & 88.7 & 88.9 & 60.1 & 41.3 & 46.8 \\
VideoChat2 \citep{li2024mvbench}    & 54.1 & 70.0 & 49.1 & 91.7 & 91.9 & 62.1 & 42.7 & 48.4 \\
Video-ChatGPT \citep{maaz2023video} & 49.3 & 64.9 & 35.2 & 65.2 & 65.9 & 44.5 & 30.6 & 34.7 \\
Video-LLaVA \citep{lin2023video}    & 59.2 & 70.7 & 45.3 & 84.3 & 84.8 & 57.3 & 39.4 & 44.7 \\
GPT4Video \citep{wang2024gpt4video} & 49.8 & 66.3 & 48.7 & 90.9 & 91.1 & 61.6 & 42.3 & 48.0 \\
7B-PLLaVA \citep{xu2024pllava}      & 62.0 & 76.6 & 56.3 & 95.1 & 85.4 & 71.2 & 49.0 & 55.5 \\
13B-PLLaVA \citep{xu2024pllava}     & 63.2 & 75.7 & 56.3 & 95.4 & 86.7 & 72.9 & 49.5 & 58.6 \\
ST-LLM \citep{liu2024st}            & 63.2 & 74.6 & 50.9 & 95.1 & 95.3 & 64.4 & 44.3 & 50.2 \\
Chat-UniVi \citep{jin2024chat}      & 54.6 & 65.0 & 45.8 & 84.5 & 85.7 & 57.9 & 39.8 & 45.2 \\
\midrule
\underline{7B-LVLM (Ours)}          & \underline{65.0} & \underline{78.1} & \underline{57.3} & \underline{97.0} & \underline{96.6} & \underline{74.5} & \underline{50.9} & \underline{59.0} \\
\rowcolor{GrayRow}
\textbf{13B-LVLM (Ours)}            & \textbf{66.7} & \textbf{79.5} & \textbf{58.3} & \textbf{98.5} & \textbf{97.8} & \textbf{76.1} & \textbf{52.3} & \textbf{59.4} \\
\bottomrule
\end{tabular}}
\vspace{-10pt}
\end{table}}
\vspace{-5pt}
\section{Conclusion}
\label{sect:conclusion}
\vspace{-5pt}
In this work, we highlight the critical role of temporal modeling in the design of modern Large Vision-Language Models (LVLMs). Throughout extensive investigation, we discover that key components, including query transformer (Q-Former), temporal-oriented training schemes, memory bank, and MoE augmentation for Q-Former, are pivotal for effective video understanding with LVLMs. Our empirical findings culminate in a step-by-step, temporal-oriented recipe for constructing effective temporal modeling capacity in LVLM. Compared with existing LVLMs, our proposed approach achieves superior performance across a broad range of standard video understanding datasets. Notably, the benefits of our recipe become more pronounced for larger-scale LVLMs, underscoring the potential of explicitly incorporating temporal modeling into large-scale architectures. 

\bibliographystyle{plainnat}
\bibliography{neurips_2025}

\newpage

\appendix
\section{Implementation Details}
\label{app:implementation_details}
\subsection*{Video Question Answering}
We formulate VideoQA as text generation task. After conducting the temporal-oriented training stage, we fine-tune the model to optimize its performance on each downstream dataset, using the averaged cross-entropy loss of each token between the generated answer and the groundtruth answer. 

\subsection*{Video Captioning}
Since the nature of the task is inherently text generation, the only remaining concern is the evaluation protocol. Because strictly adhering to surface words might not adequately assess model quality, we follow existing LVLM works \citep{lin2023video, maaz2023video} to use \textit{gpt-3.5-turbo} to judge the quality of the answer. 

\section{Dataset Descriptions}
\label{app:dataset_descriptions}
\subsection*{Temporal-Oriented Training} We conduct an additional temporal-oriented training stage after the model has been pretrained and instruction-tuned in previous works. The datasets we use consist of  InternVid \citep{wang2023internvid} and VIDAL-10M \citep{zhu2023languagebind}.
\begin{itemize}
    \item \textbf{InternVid (745K):} the original dataset comprises 234M video clips accompanied by detailed descriptions from 7M videos. Due to computational and storage limit, we use only 745K clips to train our model.
    \item \textbf{VIDAL-10M (661K):} consists of 10M short videos paired with corresponding descriptions. In our work, we utilize 661K videos to train our LVLM.
\end{itemize}

\subsection*{Video Question Answering} We evaluate on three short-term VideoQA datasets, \textit{i.e.} MSRVTT \citep{xu2016msr}, MSVD \citep{chen2011collecting}, and ActivityNet-QA \citep{caba2015activitynet}, and three long-term VideoQA datasets, \textit{i.e.} Breakfast \citep{kuehne2014language}, COIN \citep{tang2019coin}, and LVU \citep{wu2021towards}.
\begin{itemize}
    \item \textbf{MSRVTT} \citep{xu2016msr} composed of 10K YouTube videos, for VideoQA the dataset is formatted into 243K open-ended questions. We adopt the 149K-12K-73K train-val-test split to evaluate LVLMs.

    \item \textbf{MSVD} \citep{chen2011collecting} comprises 47K open-ended questions for 2K videos. We employ a split of 30K/6K/13K to divide the questions into training, validation, and testing sets, respectively.

    \item \textbf{ActivityNet-QA} \citep{caba2015activitynet} consists of 58K open-ended questions on 5.8K sampled videos from ActivityNet \citep{caba2015activitynet}.

    \item \textbf{Breakfast} \citep{kuehne2014language} encompasses 1.7K videos related to 10 actions for breakfast preparation. The model is asked to predict the action type in the video.

    \item \textbf{COIN} \citep{tang2019coin} includes 12K videos from YouTube, covering 180 diverse tasks in 12 domains related to daily life. The model is tasked with predicting the task type conducted in the video.

    \item \textbf{LVU} \citep{wu2021towards} consists of 30K videos sourced from 3K movies. Given a video, we train/test the model to predict the relationship, speaking style, scene, director, genre, writer, and release year of the video.
\end{itemize}

\subsection*{Video Captioning}
We evaluate our model on two prevalently used datasets, \textit{i.e.} MSRVTT \citep{xu2016msr} and MSVD \citep{chen2011collecting}. 
\begin{itemize}
    \item \textbf{MSRVTT} \citep{xu2016msr} consists of 200K videos paired with respective captions. To ensure fair comparison with previous works \citep{li2024llama, nguyen2024meta, he2024ma}, we use a split ratio of 130K/10K/60K for training, validation, and testing.
    \item \textbf{MSVD} \citep{chen2011collecting} contains 81K videos and corresponding captions. Following recent works \citep{li2024llama, nguyen2024meta, he2024ma}, we adopt a ratio of 49K/4K/28K to split these samples into training, validation, and testing sets.
\end{itemize}

\end{document}